\documentclass[10pt,twocolumn,letterpaper]{article}

\usepackage{iccv}              

\input{preamble}

\definecolor{iccvblue}{rgb}{0.21,0.49,0.74}
\usepackage[pagebackref,breaklinks,colorlinks,allcolors=iccvblue]{hyperref}

\def\paperID{4103} 
\def\confName{ICCV}
\def\confYear{2025}

\title{InsViE-1M: Effective Instruction-based Video Editing with \\ Elaborate Dataset Construction}

\author{
Yuhui Wu$^{1,2}$, Liyi Chen$^{1}$, Ruibin Li$^{1}$, Shihao Wang$^{1}$, Chenxi Xie$^{1,2}$, Lei Zhang$^{1,2}$\thanks{Corresponding author. This research is supported by the PolyU-OPPO Joint Innovative Research Center.}\\
{$^{1}$The Hong Kong Polytechnic University \quad
$^{2}$OPPO Research Institute} \\
{\tt\small \{yuhui.wu, liyi0308.chen, ruibin.li, shihaow.wang, chenxi.xie\}@connect.polyu.hk} \\
{\tt\small cslzhang@comp.polyu.edu.hk} \\
}

\begin{document}

\maketitle
\vspace{-2em}

\vspace{-5mm}
\begin{abstract}
   \vspace{-7mm}

   Instruction-based video editing allows effective and interactive editing of videos using only instructions without extra inputs such as masks or attributes. However, collecting high-quality training triplets (source video, edited video, instruction) is a challenging task. 
   Existing datasets mostly consist of low-resolution, short duration, and limited amount of source videos with unsatisfactory editing quality, limiting the performance of trained editing models. 
   In this work, we present a high-quality \textbf{Ins}truction-based \textbf{Vi}deo \textbf{E}diting dataset with \textbf{1M} triplets, namely \textbf{InsViE-1M}. 
   We first curate high-resolution and high-quality source videos and images, then design an effective editing-filtering pipeline to construct high-quality editing triplets for model training. 
   For a source video, we generate multiple edited samples of its first frame with different intensities of classifier-free guidance, which are automatically filtered by GPT-4o with carefully crafted guidelines. 
   The edited first frame is propagated to subsequent frames to produce the edited video, followed by another round of filtering for frame quality and motion evaluation. 
   We also generate and filter a variety of video editing triplets from high-quality images.   
   With the InsViE-1M dataset, we propose a multi-stage learning strategy to train our InsViE model, progressively enhancing its instruction following and editing ability. 
   Extensive experiments demonstrate the advantages of our InsViE-1M dataset and the trained model over state-of-the-art works. 
   Codes are available at \href{https://github.com/langmanbusi/InsViE}{InsViE}.
\end{abstract}
    
\vspace{-5mm}
\section{Introduction}
\label{sec:introduction}
\vspace{-1mm}

\noindent
Promising progress on video editing has been achieved in recent years.
Many previous works~\cite{qi2023fatezero,geyer2023tokenflow,kara2024rave,li2024vidtome,fan2024videoshop,ku2024anyv2v,cohen2024slicedit} are training-free methods based on pre-trained image/video generation models~\cite{Rombach_2022_CVPRsd,blattmann2023stablesvd}, yet they are limited in generalization ability, long time consumption and the need of paired captions.  
One-shot methods~\cite{wu2023tuneavideo,ouyang2024i2vedit,gu2024videoswap} improve the temporal consistency by over-fitting on each single video, but at the price of more time consumption.
Training-based methods perform video editing by introducing masks~\cite{li2025magiceraser,hu2024vivid}, edited first-frame~\cite{liu2024generativegenprop}, or instructions~\cite{brooks2023instructpix2pix,zhang2024effived,cheng2023consistentinsv2v}.
Compared with other kinds of training-based video editing methods, instruction-based editing is more user-friendly as it only needs the instruction to represent the expected editing target. Inspired by the success of instruction-based image editing ~\cite{guo2024focus,huang2024smartedit}, researchers have proposed several methods~\cite{qin2024instructvid2vid,zhang2024effived,cheng2023consistentinsv2v} to construct instruction-based video editing datasets and train the models.

\begin{table}[t]
  \centering
  \caption{Statistics of instruction-based video editing datasets. 
  }
  \vspace{-3mm}
  \scalebox{.75}{
    \begin{tabular}{c||c|c|c|c|c}
    \toprule
    Datasets & Amount & Resolution & Frames & Real & Filter \\
    \midrule
    InstructVid2Vid~\cite{qin2024instructvid2vid} &
    N/A & N/A & N/A & $\checkmark$ & $\times$ \\
    EffiVED~\cite{zhang2024effived} &
    155k & 512$\times$512 & 8 & $\checkmark$ & $\times$ \\
    InsV2V~\cite{cheng2023consistentinsv2v} &
    404k & 256$\times$256 & 16 & $\times$ & $\times$ \\
    \textbf{Ours} &
    1M & 1024$\times$576 & 25 & $\checkmark$ & $\checkmark$ \\
    \bottomrule
    \end{tabular}}
  \label{tab:dataset_comparison}%
  \vspace{-6mm}
\end{table}%

\begin{figure*}[t]
  \centering
   \includegraphics[width=\linewidth]{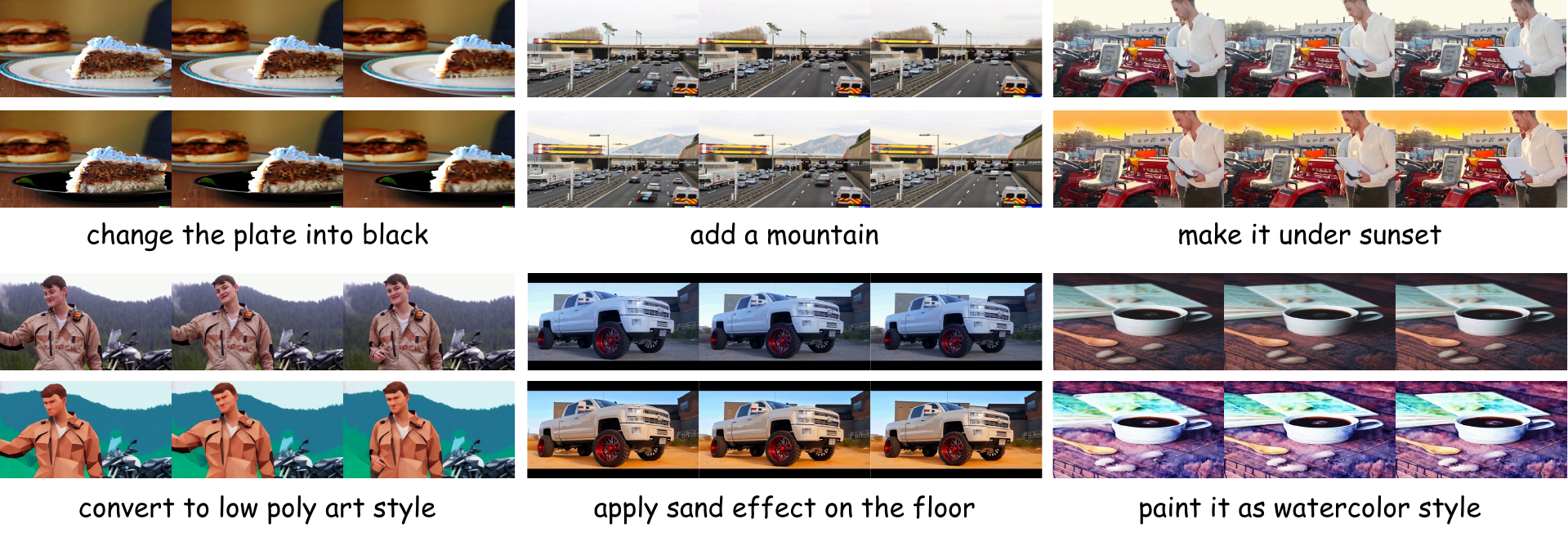}
   \setlength{\abovecaptionskip}{-0.5cm}
   \setlength{\belowcaptionskip}{-0.5cm}
   \caption{
   Sample triplets of our InsViE-1M dataset. For each sample, from top to bottom: original video, edited video, instruction. 
   }
   \label{fig:samples_visual}
   \vspace{-2mm}
\end{figure*}

However, the training data construction for instruction-based video editing (InsViE) is much more challenging than that for image editing. 
InstructVid2Vid~\cite{qin2024instructvid2vid} and EffiVED~\cite{zhang2024effived} create edited videos utilizing one-shot methods \cite{wu2023tuneavideo,ouyang2024codef}, which is highly time consuming. As a result, only 24K videos with 8 frames per video of 256$\times$256 resolution are provided in ~\cite{zhang2024effived}, whose video quantity and quality are not sufficient to train a robust video editor. 
InsV2V~\cite{cheng2023consistentinsv2v} synthesizes 400K editing pairs; however, the source videos are synthetic, and hence the trained models are less effective when editing real-world videos. In addition, the limited video resolution (256$\times$256) of this dataset limits the application of the trained models. As summarized in~\cref{tab:dataset_comparison}, existing InsViE datasets suffer from low-resolution, short video duration, and insufficient quantity and quality, making InsViE remain a challenging task.

To address the limitation, we propose to construct \textbf{InsViE-1M}, an instruction-based video editing dataset, including 1M high-quality training triplets (source video, edited video, instruction). 
Specifically, we first collect a large amount of 1080p real-world source videos and utilize large vision-language models (VLMs) \cite{hong2024cogagent} to generate captions and instructions, then present an effective pipeline to generate high-quality edited videos. 
Previous data construction pipelines commonly employ random parameters to synthesize edited videos~\cite{zhang2024effived,cheng2023consistentinsv2v}, which is difficult to ensure the editing quality.
Instead, we propose a two-stage editing-filtering pipeline for video editing triplet generation. 
In the first stage, we employ a powerful image editing model to edit the first frame of each video, and output multiple edited samples by varying the classifier-free guidance (CFG) within a range. 
The edited samples are then examined by GPT-4o~\cite{gpt4o} with our carefully prepared screening guideline, aiming to find the best candidate from a comprehensive set of evaluation metrics.
In the second stage, we propagate the edited first frame to subsequent frames and apply another round of filtering based on frame quality and motion consistency. 
We design a scoring guideline for GPT-4o to evaluate the corresponding frames for each video before and after editing, and adopt optical flow endpoint error to evaluate motion consistency.
In addition, we generate a number of source/edited videos from high-quality source/edited image pairs, as well as a set of static videos from high-quality images as parts of InsViE-1M. 
~\cref{fig:samples_visual} illustrates some sample triplets of our InsViE-1M dataset, including the source videos, their edited counterparts and the associated instructions. 

Most previous InsViE models are built upon image models~\cite{qin2024instructvid2vid,cheng2023consistentinsv2v}, which often introduce flicker artifacts due to poor motion consistency. With our established InsViE-1M dataset, we employ pre-trained video generation models \cite{yang2024cogvideox} and propose a multi-stage learning strategy to progressively train an InsViE model to enhance its editing ability. 
In addition, we adopt LPIPS loss as a complement of $L_2$ loss for detail preservation, avoiding the decay of editing effect in the propagation process of subsequent frames. 

Our contributions can be summarized as follows. First, we construct a high-quality instruction-based video editing dataset, \ie, InsViE-1M, through an elaborately designed two-stage editing-filtering pipeline. Second, we present an InsViE model, which is the first built upon video generation models. 
Experiments demonstrate the advantages of our InsViE-1M dataset and the trained InsViE model. 

\vspace{-3mm}
\section{Related Work}
\label{sec:relatedwork}
\vspace{-2mm}

\begin{figure*}[t]
  \centering
   \includegraphics[width=\linewidth]{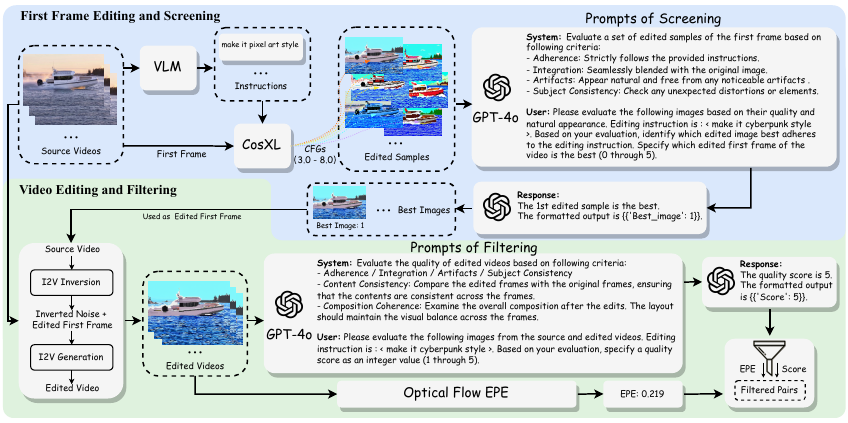}
   \setlength{\abovecaptionskip}{-0.4cm}
   \setlength{\belowcaptionskip}{-0.4cm}
   \caption{
   Our proposed two-stage editing-filtering pipeline for generating triplets from real-world videos. 
   In the $1^{st}$ stage, we edit the first frame of each video and screen the best edited sample. 
   In the $2^{nd}$ stage, we edit videos and filter them using GPT-4o and optical flow.
   }
   \label{fig:data_construction}
   \vspace{-2mm}
\end{figure*}

\textbf{Training-free video editing}.
Many existing video editing methods rely on pre-trained large image/video models~\cite{qi2023fatezero,geyer2023tokenflow,li2024vidtome,kara2024rave,cong2023flatten,cohen2024slicedit,ku2024anyv2v,fan2024videoshop} to perform DDIM~\cite{song2020denoisingddim} inversion and denoising without training on any data.  
FateZero~\cite{qi2023fatezero} introduces attention blending to disentangle the editing of objects and background. To ensure motion consistency, FLATTEN~\cite{cong2023flatten} and RAVE~\cite{kara2024rave} use optical flow and random noise shuffling to improve the editing effect. 
Tokenflow~\cite{geyer2023tokenflow} and VidToMe~\cite{li2024vidtome} merge the intermediate tokens to accelerate the editing process.
AnyV2V~\cite{ku2024anyv2v} and Videoshop~\cite{fan2024videoshop} adopt pre-trained image-to-video (I2V) models~\cite{blattmann2023stablesvd,zhang2023i2vgen} to edit the whole video with a given edited first-frame, providing more interaction with users. 
However, training-free methods often yield unsatisfactory results due to the inherent gap between generation and editing tasks. 
Furthermore, it is difficult to ensure the quality of the generated samples when we construct large-scale datasets directly using training-free methods. Therefore, we design an elaborate pipeline to filter both the edited first frame and the final edited video, which significantly improves the quality of our dataset. 

\textbf{One-shot video editing}. 
One-shot editing methods~\cite{wu2023tuneavideo,gu2024videoswap,ouyang2024codef,ouyang2024i2vedit} over-fit on a single video to yield better visual effects. 
Tune-A-Video~\cite{wu2023tuneavideo} learns the motion of each video by updating spatial and temporal attention. 
VideoSwap~\cite{gu2024videoswap} introduces semantic points optimized on source video frames to manipulate the object and preserve the background. 
CoDeF~\cite{ouyang2024codef} is trained to extract the canonical and deformation fields from each video to render the edited video. 
I2VEdit~\cite{ouyang2024i2vedit} first extracts the coarse motion of the given video by trainable LoRA~\cite{hu2021lora}, then uses attention matching to refine the appearance in the inference stage. 
Although these methods generate videos with better motion consistency, the time consumption is considerable due to the online optimization process for each video.

\textbf{Training-based video editing}.
Recent works~\cite{brooks2023instructpix2pix,cheng2023consistentinsv2v,zhang2024effived,hu2024vivid,liu2024generativegenprop} have started to build large video editing datasets to train video editing models. 
ViViD-10M~\cite{hu2024vivid} and GenProp~\cite{liu2024generativegenprop} utilize mask and the first frame to edit the video by pre-trained large models. As initiated by the work of 
InstructVid2Vid~\cite{qin2024instructvid2vid}, a class of methods~\cite{brooks2023instructpix2pix,cheng2023consistentinsv2v,zhang2024effived} adopt the instruction-based editing approach. 
For more robust performance, InsV2V~\cite{cheng2023consistentinsv2v} creates a large dataset with more than 400k synthetic pairs and trains the editing model by introducing temporal layers to the model of InstructPix2Pix~\cite{brooks2023instructpix2pix}. 
Since image editing models often fail to preserve temporal consistency, EffiVED~\cite{zhang2024effived} generates 155k training pairs, including 24k real-world source videos, to train a video editing model upon video generation model. 
Senorita-2M~\cite{zi2025senorita} builds multiple expert models to create high quality editing effects, creating 2M samples.
However, fine-tuning a pre-trained video generation model requires extensive high-quality triplets (source video, edited video, instruction) to achieve sufficient instruction-based editing capability. 
In this paper, we construct a large video editing dataset for training instruction-based editing models by collecting high-quality source videos and designing effective filtering strategies to improve training data quality. 
Compared with Senorita-2M that uses trained classifier and CLIP for filtering, we apply multi-stage filtering by GPT-4o. Actually, these strategies can be integrated to further improve the quality of data.

\vspace{-3mm}
\section{InsViE-1M Dataset Consruction}
\label{sec:insvie-1m}
\vspace{-1mm}

\noindent
Some recent works ~\cite{qin2024instructvid2vid,cheng2023consistentinsv2v,zhang2024effived} have been proposed to construct datasets for training instruction-based video editing models, yet their data quality and quantity remain limited.
Therefore, we propose InsViE-1M, a large-scale instruction-based video editing dataset consisting of 1M high-quality triplets (source video, edited video, instruction).
We collect high-resolution video clips and images from publicly available sources~\cite{nan2024openvid,pexels, brooks2023instructpix2pix,zhang2024magicbrush}, and develop a two-stage pipeline to generate and screen high-quality triplets. These triplets are generated from three types of source data: real-world videos, realistic videos synthesized from image editing pairs, and static videos converted from real-world images, as detailed below. 

\vspace{-2mm}
\subsection{Triplet generation from real-world videos}
\label{subsec:realdata}
\vspace{-1mm}
We curate high-quality videos from Pexel~\cite{pexels} and Openvid~\cite{nan2024openvid}, and design an elaborate two-stage pipeline to generate and screen the edited samples. As shown in~\cref{fig:data_construction}, we perform first frame editing and screening in the $1^{st}$ stage, and video editing and filtering in the $2^{nd}$ stage.

\begin{figure}[t]
  \centering
   \includegraphics[width=\linewidth]{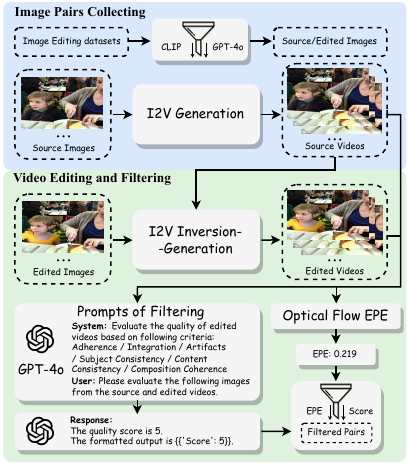}
   \setlength{\abovecaptionskip}{-0.4cm}
   \setlength{\belowcaptionskip}{-0.4cm}
   \caption{
   Overview of the proposed two-stage editing-filtering pipeline for generating triplets from image editing pairs. 
   Source/edited images are curated from image editing datasets. 
   We edit and filter the videos using GPT-4o and optical flow.
   }
   \label{fig:data_construction_detail}
   \vspace{-3mm}
\end{figure}

\textbf{First frame editing and screening.}
Existing real-world video datasets \cite{nan2024openvid,pexels} contain videos and corresponding captions, while the captions are either too long or only a few words, unsuitable for instruction-based editing. 
We re-caption the videos and produce various instructions using off-the-shelf large language models~\cite{hong2024cogagent}. 
Specifically, we generate captions of several keyframes and summarize them into the final caption, which includes the relationships and characteristics of objects. 
The instructions are then generated based on the refined captions. The details of this process can be found at the \textbf{supplementary file}. 

With the instruction available, one may employ existing video editing methods to generate edited videos.
However, one-shot methods suffer from high time consumption, while training-free methods cannot generalize well to large-scale real-world videos. 
Recent I2V-model-based editing methods~\cite{ku2024anyv2v,fan2024videoshop} can produce videos with fine details and consistent motion due to the use of video generation models like SVD~\cite{blattmann2023stablesvd}.
These methods rely on the quality of the edited first frame, which can be obtained by image editing models such as CosXL~\cite{cosxl}. In addition,
these generative image editing models can produce different types of edits with different levels of classifier-free-guidance (CFG).
Thus, we generate multiple edited samples by using different CFGs, instead of employing a random parameter to produce a single output~\cite{cheng2023consistentinsv2v,zhang2024effived}. Based on our experience, most of the best samples can be generated with CFGs from 3.0 to 8.0. Therefore, we set CFG within [3, 8] to generate 6 edited samples.
For the first frame of each video, the edited samples and the instruction are input to GPT-4o~\cite{gpt4o} to select the best result.
As shown in~\cref{fig:data_construction}, we conduct the selection from four aspects: adherence to instruction, natural integration of edits, absence of artifacts, and subject matter consistency.
Through repetitive generation and evaluation, we find the best edit of the first frame for each video.

\textbf{Video editing and filtering.}
In this stage, we propagate the edited first frame to subsequent frames and filter the edited videos for training. The purpose of automated evaluation at this stage is to filter out videos with unsatisfactory editing effects, instead of selecting the best.
Our propagation pipeline adopts SVD~\cite{blattmann2023stablesvd} as the base model.
Similar to previous training-free methods~\cite{ku2024anyv2v,fan2024videoshop}, we first invert the source video into noise and then generate the edited video from inverted noise using the best edited image as conditional input.
We only generate one result for each video due to the significant time consumption of video generation.

We extract three frames respectively from source/edited videos as samples to be evaluated by GPT-4o~\cite{gpt4o}.
The evaluation is conducted from six perspectives. In addition to the four used in stage 1, in this stage we further consider composition coherence across frames and content consistency across frames.
In this way, we obtain the score (ranging from 1 to 5) for each video by setting tailored user prompts, as shown in~\cref{fig:data_construction}.
Furthermore, we complement the motion evaluation on discrete frames by calculating the optical flow of the whole video.
We employ GMFlow~\cite{xu2022gmflow} to compute the optical flow of each video pair and measure the endpoint error (EPE), and filter out the videos with low evaluation scores and high EPE values. 
Finally, an amount of 450k triplets are created from real-world videos.

\begin{table}[t]
  \centering
  \caption{Statistics of InsViE-1M dataset. 
  }
  \vspace{-3mm}
  \scalebox{.72}{
    \begin{tabular}{c||c|c|c|c|c|c}
    \toprule
    Data Source & Amount & Src Res & Out Res & Frames & GPT & EPE \\
    \midrule
    Real videos &
    450,790     & 1080p          & 1024$\times$576 & 25 & 3.47 & 0.79 \\
    Image pairs &
    110,374     & 512$\times$512 & 1024$\times$576 & 25 & 3.43 & 1.22 \\
    Real images &
    458,429     & 1080p          & 1024$\times$576 & 25 & 4.07 & 0.16 \\
    Overall     &
    1,019,593   & -              & 1024$\times$576 & 25 & 3.74 & 0.55\\
    \bottomrule
    \end{tabular}}
  \label{tab:dataset_statistic}%
  \vspace{-5mm}
\end{table}%

\vspace{-1mm}
\subsection{Triplet generation from image editing pairs}
\label{subsec:imagedata}
\vspace{-1mm}
Existing image editing datasets~\cite{brooks2023instructpix2pix,zhang2024magicbrush} possess a variety of image pairs with diverse editing types, which can be employed to generate video editing data.
We first select 150k image editing triplets from InstructPix2Pix~\cite{brooks2023instructpix2pix} based on metrics such as CLIP similarity, CLIP directional score, and the GPT-4o score we used in~\cref{subsec:realdata}. We then employ all the 10k image editing pairs in  MagicBrush~\cite{zhang2024magicbrush} since this dataset is manually annotated with precise editing effects.
As shown in~\cref{fig:data_construction_detail}, with these image editing pairs, we first generate the source videos from the source images through I2V generation model SVD~\cite{blattmann2023stablesvd}.
Then, we perform video editing and filtering using the same process as the $2^{nd}$ stage used in~\cref{subsec:realdata}. Finally, we generate 110k filtered video editing triplets from image editing pairs. 

\vspace{-1mm}
\subsection{Generate static video triplets from images}
\label{subsec:staticvideo}
\vspace{-1mm}

\begin{figure*}[t]
  \centering
   \includegraphics[width=\linewidth]{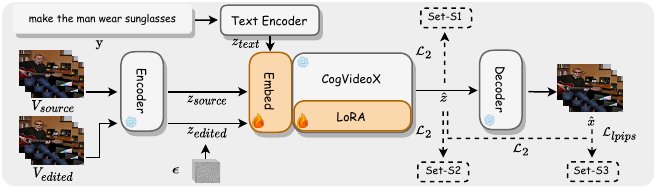}
   \setlength{\abovecaptionskip}{-0.3cm}
   \setlength{\belowcaptionskip}{-0.3cm}
   \caption{
   Training framework of our InsViE model.
   }
   \label{fig:model}
   \vspace{-3mm}
\end{figure*}

Since producing video editing data is costly, previous works \cite{hu2024vivid,zhang2024effived,tu2025videoanydoor} have proposed to utilize images as part of data for mixed training. 
However, directly training on image data may not align well with real-world videos, hindering the training of video editing models. Considering the fact that a portion of real-world videos feature only on camera movement, we translate the images into static videos with manual camera operations, including (1) zoom-in, (2) zoom-out, (3) move left, (4) move right, (5) move down, (6) move up.
The edited image is generated by using CosXL~\cite{cosxl} as in the $1^{st}$ stage of~\cref{subsec:realdata}.
We crop each source and edited image pair at equal intervals to create image sequence pair, which is then processed with bilinear interpolation to enhance the video smoothness. 
Then we apply the filtering process as in the $2^{nd}$ stage of~\cref{subsec:realdata}.
In total, we produce 450k triplets of static videos that only possess camera movement for training.

In~\cref{tab:dataset_statistic}, we summarize the composition and statistics of our InsViE-1M dataset.
``Src Res'' and ``Out Res'' denote the source resolution and output resolution. ``GPT'' and ``EPE'' are the averages of GPT scores and optical flow EPE values. Examples of the triplets construction process can be found in the \textbf{supplementary file}. 

\vspace{-2mm}
\section{InsViE Model Training}
\label{sec:insviemodel}
\vspace{-1mm}

In this section, we train the instruction-based editing model by finetuning CogVideoX-2B~\cite{yang2024cogvideox} on the InsViE-1M dataset. 
Instruction-based video editing aims to edit videos according to the given instructions, which can be achieved through a conditional diffusion process. 
Specifically, the source video $V_{source}$, the text instruction $y$, and the edited video $V_{edited}$ are given in the training phase. 
We first input the video pairs to the visual encoder $\mathcal{E}$ and extract the latents $z_{source} = \mathcal{E}(V_{source})$ and $z_{edited} = \mathcal{E}(V_{edited})$. 
The instruction is fed to the text encoder to produce the text condition $z_{text}$. 
In the diffusion process, noise $\epsilon$ is added to $z_{edited}$ to generate latent noise $z_t$ over timesteps $t \in T $. 
The editing model predicts the noise added to the noisy latent $z_t$ by taking $z_{source}$ and $z_{text}$ as conditions. 
The objective function of the latent diffusion process is: 
\vspace{-0.2cm}
\begin{align}
    \mathcal{L} &= \mathbb{E}_{\epsilon \sim \mathcal{N}(0,1), t}\left[\| \epsilon - \epsilon_\theta(t, z_t, z_{source}, z_{text}) \|_2^2\right], 
    \label{eq:diffusion}
    \vspace{-0.2cm}
\end{align}
where $\epsilon_\theta$ refers to the denoising network. 

\vspace{-1mm}
\subsection{Model architecture}
\vspace{-1mm}
Unlike methods that finetune image generation models~\cite{qin2024instructvid2vid,cheng2023consistentinsv2v}, we develop the InsViE model by finetuning the video generation model CogVideoX~\cite{yang2024cogvideox} to enhance motion consistency.
As illustrated in~\cref{fig:model}, we begin by encoding the source video $V_{source}$ and the instruction $y$ into latent representations $z_{source}$ and $z_{text}$, which are then concatenated and passed through an embedding layer.
Then, the output of embedding layer and  $z_{text}$ are concatenated and fed into DiT to produce the denoised latent $\hat{z}$.
Finally, InsViE model outputs $\hat{x}$ by decoding $\hat{z}$ as $\hat{x} = \mathcal{D}(\hat{z})$, where $\mathcal{D}$ is the decoder. 
During the training phase, both the embedding layer and LoRA~\cite{hu2021lora} parameters are set to be trainable.

\vspace{-1mm}
\subsection{Training and sampling}
\label{subsec:training}
\vspace{-1mm}

\begin{table}[t]
  \centering
  \caption{Dataset settings of multi-stage training. 
  }
  \vspace{-3mm}
  \scalebox{.8}{
    \begin{tabular}{c||c|c|c|c}
    \toprule
    \; Subset \; & \; Amount \; & Static:Real & \; GPT \; & \; EPE \;  \\
    \midrule
    Set-S1 &
    1,019,593   & 1:1 & 3.74 & 0.55 \\
    Set-S2 &
    226,772     & 1:1 & 4.34 & 0.42 \\
    Set-S3 &
    680,316     & 5:1 & 4.41 & 0.29 \\
    \bottomrule
    \end{tabular}}
  \label{tab:dataset_training}%
  \vspace{-6mm}
\end{table}%

\noindent
\textbf{Multi-stage training.} 
Inspired by the training of video generation models~\cite{yang2024cogvideox,zheng2024opensora}, we implement the model training in a multi-stage manner to progressively enhance our model's editing capability.
As mentioned in~\cref{sec:insvie-1m}, we designed an evaluation guideline for GPT-4o to score each video and filter the dataset.
In the \textbf{first} stage, we train our InsViE model on the filtered dataset, referred to as \textbf{Set-S1}, to learn the general ability of instruction-based editing.
The $L_2$ loss is applied to the output latent $\hat{z}$, as shown in~\cref{fig:model}.
In the \textbf{second} stage, we select video pairs with higher GPT-4o scores and lower EPE values to create a new training dataset, referred to as \textbf{Set-S2}, to enhance the editing quality. 
The $L_2$ loss on the output latent is still used in this stage.
Then, we focus on enhancing the video fidelity in the \textbf{final} stage.
We augment \textbf{Set-S2} with more static video pairs to construct \textbf{Set-S3}, as the static images we collected are of higher visual quality. The LPIPS loss on image domain data $\hat{x}$ is also used to encourage the model outputting videos with improved naturalness.
The detailed composition of the subsets of InsViE-1M are illustrated in~\cref{tab:dataset_training}.

\noindent
\textbf{Sampling.}
For sampling, we follow InstructPix2Pix~\cite{brooks2023instructpix2pix} and employ different CFGs for vision and text conditions: 
\vspace{-0.15cm}
\begin{align}
\hat{\epsilon}_{\theta}(z_t, & z_{source}, z_{text}) = \epsilon_{\theta}(z_t, \varnothing, \varnothing) \notag \\
& \quad + s_V \cdot \left( \epsilon_{\theta}(z_t, z_{source}, \varnothing) - \epsilon_{\theta}(z_t, \varnothing, \varnothing) \right) \\
& \quad + s_T \cdot \left( \epsilon_{\theta}(z_t, z_{source}, z_{text}) - \epsilon_{\theta}(z_t, \varnothing, \varnothing) \right) \notag
\vspace{-0.15cm}
\end{align}
where $s_V$ and $s_T$ denote the CFG scales of input video and instruction, respectively. During training, we randomly drop the $z_{source}$ and $z_{text}$ for 5\% of examples to align with the dual-condition CFG.

\vspace{-1mm}
\section{Experiment}
\label{sec:experiment}

\vspace{-1mm}
\subsection{Experimental Settings}
\label{subsec:experimentsetting}
\vspace{-1mm}
\noindent
\textbf{Implement details.}
Our InsViE model is finetuned on the CogVideoX-2B model~\cite{yang2024cogvideox} using LoRA~\cite{hu2021lora} with rank of 128.  
The training resolution is set to $720\times480$ by random cropping on the $1024\times576$ resolution source/edited videos in our InsViE-1M dataset. 
The training is conducted on 64 NVIDIA A100 GPUs for 40k iterations. The batch-size is 128. To be specific, we train our model on Set-S1, Set-S2 and Set-S3 for 20k, 10k, 10k iterations, respectively. 
Overall, it takes about 90s/20G/1GPU and 100h/75G/64GPUs for inference and training.

\noindent
\textbf{Compared methods.}
We compare our model with several representative methods, including SD-based training-free methods FateZero~\cite{qi2023fatezero}, TokenFlow~\cite{geyer2023tokenflow} and RAVE~\cite{kara2024rave}, SVD-based training-free method Videoshop~\cite{fan2024videoshop}, and training-based method InsV2V~\cite{cheng2023consistentinsv2v}, and EVE~\cite{singer2024tgveplus}. 

\begin{figure*}[t]
  \centering
   \includegraphics[width=\linewidth]{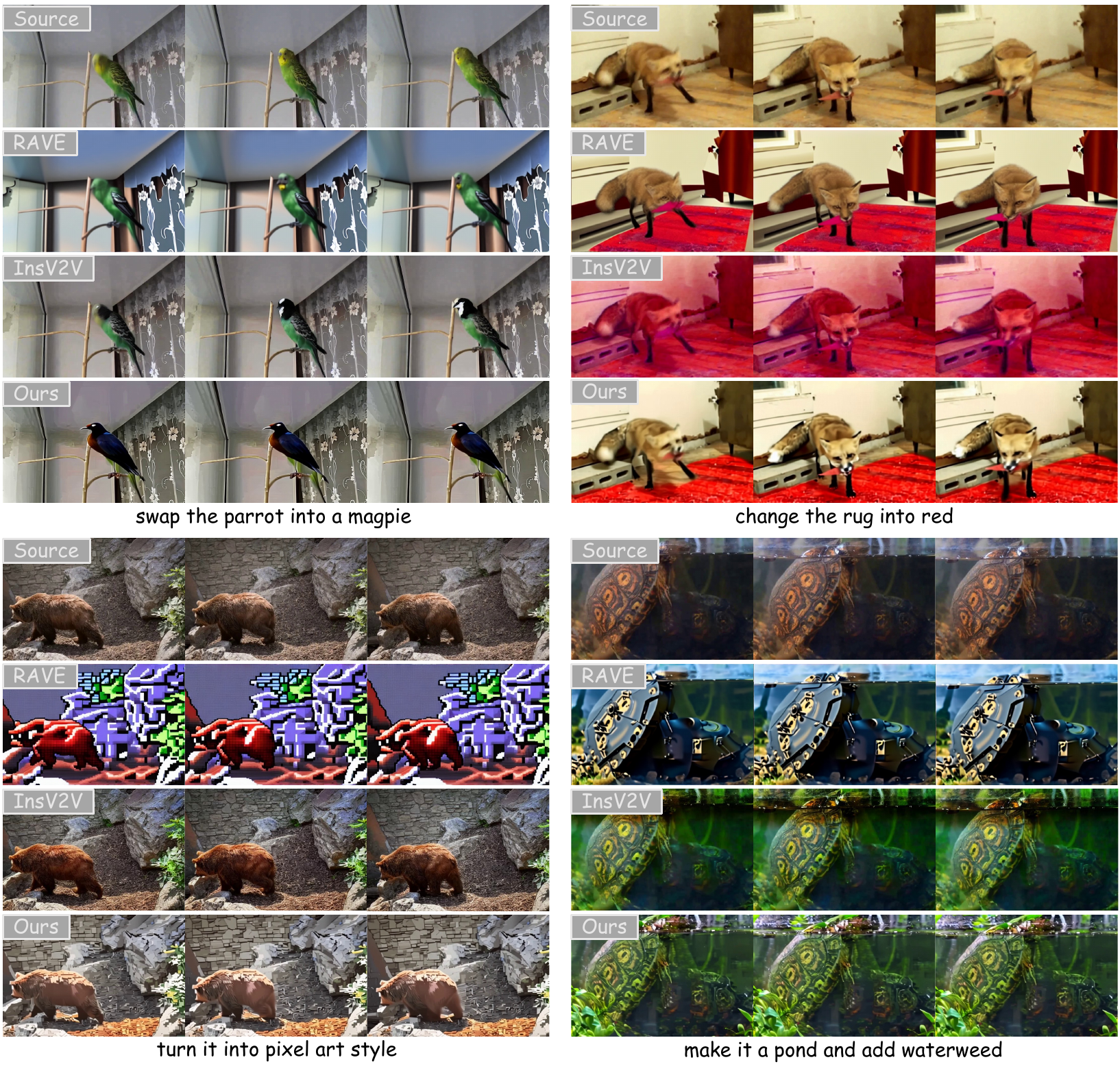}
   \setlength{\abovecaptionskip}{-0.3cm}
   \setlength{\belowcaptionskip}{-0.4cm}
   \vspace{-2mm}
   \caption{
   Visual comparison between our InsViE model and state-of-the-art methods.
   }
   \label{fig:exp_1}
   \vspace{-3mm}
\end{figure*}

\noindent
\textbf{Evaluation metrics.}
Following previous methods and the benchmark in awesome-diffusion-v2v~\cite{sun2024diffusioneditingsurvey}, we evaluate the video editing result from three aspects. 
(1) Temporal consistency (TC). We use CLIP~\cite{radford2021learningclip} temporal score to compute the CLIP embedding differences between frames, and use optical flow EPE ~\cite{xu2022gmflow} to assess the motion smoothness. 
(2) Textual alignment (TA). We employ the CLIP text-image embedding similarity and the pick score, which represents well human perception, for assessing textual alignment. 
(3) Video quality (VQ). We employ DOVER~\cite{wu2023exploringdover}, which is a state-of-the-art video quality assessment method trained on human-ranked video datasets, for evaluating edited video quality.
In addition, for each of the three aspects mentioned above, we provide (4) the average score of GPT-4o~\cite{gpt4o}, which is introduced in the scoring procedure in our data filtering stage (ranging from 1 to 5), to more comprehensively evaluate the quality of edited videos by competing methods.
For TGVE/TGVE+, we use the same metrics in EVE~\cite{singer2024tgveplus}, such as $\mathrm{ViCLIP_{\mathit{dir}}}$ and $\mathrm{ViCLIP_{\mathit{out}}}$.

\noindent
\textbf{Testing dataset.} As in previous methods \cite{geyer2023tokenflow,fan2024videoshop,kara2024rave}, we collect 100 video samples from DAVIS~\cite{davis2017pont}, YoutubeVOS~\cite{xu2018youtube} and Pexels~\cite{pexels} as the test set. 
We caption the videos to generate the corresponding instructions and paired captions using large VLMs \cite{hong2024cogagent} for instruction-based and training-free methods, respectively. Considering the fact that our method can process $720\times480$ resolution video, while the competing methods can only process $512\times512$ or lower resolution video, we collect 50 videos at each of the above two resolutions to fit better previous methods. For each method, we resize the video to its default resolution before editing, then resize the edited video back to the original video resolution.
We also compare our model with baselines on TGVE~\cite{wu2023tgve} and TGVE+~\cite{singer2024tgveplus}. We conduct experiments by resizing the videos to $720\times480$ and resizing back to $480\times480$ after editing for evaluation.

\vspace{-1mm}
\subsection{Quantitative Results}
\vspace{-1mm}

We first make a comprehensive quantitative comparison between our InsViE model and state-of-the-art video editing methods. As shown in~\cref{tab:benchmark_comparison}, we evaluate the edited videos in three aspects with a total of eight metrics. One can see that our method achieves the best results in all metrics. 
In terms of temporal consistency, our method achieves a CLIP score of 0.956, indicating its superior ability to maintain coherence across video frames. 
Meanwhile, InsViE attains the lowest optical flow EPE, confirming its effectiveness in preserving motion dynamics and seamless transitions. 
Notably, the trend of GPT temporal score is not entirely consistent with EPE, indicating that GPT-4o cannot fully capture the video motion dynamics using the partially sampled frames. 
Therefore, incorporating optical flow into the filtering process is essential.
In terms of textual alignment, InsViE achieves significantly higher CLIP score than its competitors, demonstrating strong capability in aligning visual content with textual descriptions. The Pick score and GPT score show similar trend, further underscoring the accuracy of InsViE in generating contextually appropriate content. 
Finally, InsViE surpasses previous methods on GPT quality score and DOVER score, confirming the high visual quality of produced videos. 
We can see similar trends on the TGVE/TGVE+ datasets in~\cref{tab:benchmark_comparison_tgve} . We compare our method with TokenFlow, InsV2V and EVE, which are the best performers on TGVE.
Overall, these quantitative results demonstrate InsViE's comprehensive advantages in instruction-based editing over existing methods.

\vspace{-1mm}
\subsection{Visual Comparisons}
\vspace{-1mm}

We then provide visual comparisons between our InsViE and its competitors in ~\cref{fig:exp_1}. Due to page limit, we compare with representative methods RAVE and InsV2V in the main paper, while the comparison with more methods can be found in the \textbf{supplementary file}. We can see that for the swapping task, RAVE~\cite{kara2024rave} and InsV2V~\cite{cheng2023consistentinsv2v} fail to completely swap the parrot but only edit a small part of it (\eg, RAVE changes the wing and InsV2V changes the head). In contrast, InsViE swaps the whole parrot into the magpie.
In the local color editing task, InsViE edits the rug precisely without introducing artifacts into other regions, whereas InsV2V~\cite{cheng2023consistentinsv2v} turns the whole screen red and RAVE~\cite{kara2024rave} makes the shadow of the cabinet into an unknown object.
Without the need for edited first frame or masks, instruction-based methods perform better in global editing task, while InsV2V~\cite{cheng2023consistentinsv2v} struggles to produce conspicuous effects of pixel art style. 
RAVE~\cite{kara2024rave} produces striking editing effects but destroys the original video content.
InsViE achieves a good balance between the layout and color of video.
In mixed editing task, InsViE successfully recognizes each editing key word and performs the corresponding edits in the video. 
However, InsV2V~\cite{cheng2023consistentinsv2v} omits the addition of waterweed, and RAVE~\cite{kara2024rave} generates unnatural contents. 
In short, our InsViE can edit the video with more visually pleasing effects.

\begin{table*}[ht]
  \centering
  \caption{Quantitative comparison with existing methods. OF denotes optical flow. The best results are highlighted in \textbf{bold}.
  }
 \vspace{-2mm}
    \scalebox{.87}{
    \begin{tabular}{c||c|c|c||c|c|c||c|c}
    \toprule
    \multirow{2}[2]{*}{\textbf{Methods}} 
    & \multicolumn{3}{c||}{\textbf{Temporal Consistency}} 
    & \multicolumn{3}{c||}{\textbf{Textual Alignment}} 
    & \multicolumn{2}{c}{\textbf{Video Quality}}
    \\
    \cmidrule{2-9} 
    & \textbf{CLIP} $\uparrow$ & \textbf{OF EPE} $\downarrow$ & \textbf{GPT Score} $\uparrow$
    & \textbf{CLIP} $\uparrow$ & \textbf{Pick Score} $\uparrow$ & \textbf{GPT Score} $\uparrow$ 
    & \textbf{DOVER} $\uparrow$ & \textbf{GPT Score} $\uparrow$
    \\
    \midrule
    FateZero~\cite{qi2023fatezero}
    & 0.939 & 13.07 & 2.89 & 19.27 & 18.74 & 3.62 & 0.489 & 3.23 \\
    Tokenflow~\cite{geyer2023tokenflow} 
    & 0.951 & 6.58 & 3.32 & 18.59 & 18.63 & 3.34 & 0.566 & 3.77 \\
    RAVE~\cite{kara2024rave} 
    & 0.953 & 5.18 & 3.32 & 18.72 & 18.65 & 3.35 & 0.494 & 3.29 \\
    Videoshop~\cite{fan2024videoshop} 
    & 0.952 & 5.02 & 3.81 & 18.92 & 18.78 & 3.61 & 0.501 & 3.42 \\
    InsV2V~\cite{cheng2023consistentinsv2v} 
    & 0.951 & 4.97 & 3.76 & 19.01 & 18.75 & 3.66 & 0.559 & 3.68 \\
    \textbf{Ours} 
    & \textbf{0.956} & \textbf{4.84} & \textbf{3.88} & \textbf{19.37} & \textbf{18.91} & \textbf{3.84} & \textbf{0.567} & \textbf{3.79} \\
    \bottomrule
    \end{tabular}}
  \label{tab:benchmark_comparison}%
  \vspace{-3mm}
\end{table*}%

\vspace{-1mm}
\subsection{Ablation Study}
\vspace{-1mm}

\begin{table*}[ht]
  \centering
  \caption{Ablation study on data filtering and multi-stage training, where OF denotes optical flow.
  }
  \vspace{-3mm}
    \scalebox{.85}{
    \begin{tabular}{c||c|c|c||c|c|c||c|c}
    \toprule
    \multirow{2}[2]{*}{\textbf{Training Settings}} 
    & \multicolumn{3}{c||}{\textbf{Temporal Consistency}} 
    & \multicolumn{3}{c||}{\textbf{Textual Alignment}} 
    & \multicolumn{2}{c}{\textbf{Video Quality}}
    \\
    \cmidrule{2-9} 
    & \textbf{CLIP} $\uparrow$ & \textbf{OF EPE} $\downarrow$ & \textbf{GPT Score} $\uparrow$
    & \textbf{CLIP} $\uparrow$ & \textbf{Pick Score} $\uparrow$ & \textbf{GPT Score} $\uparrow$ 
    & \textbf{DOVER} $\uparrow$ & \textbf{GPT Score} $\uparrow$
    \\
    \midrule
    w/o GPT filter
    & 0.944 & 5.66 & 3.61 & 18.47 & 18.23 & 3.41 & 0.515 & 3.57 \\
    w/o OF filter
    & 0.941 & 6.58 & 3.54 & 18.73 & 18.37 & 3.69 & 0.490 & 3.54 \\
    \midrule
    Stage 1
    & 0.950 & 5.08 & 3.76 & 18.87 & 18.40 & 3.68 & 0.510 & 3.67 \\
    Stage 1\&2
    & 0.951 & 4.87 & 3.81 & 19.03 & 18.65 & 3.77 & 0.519 & 3.68 \\
    Stage 1\&2\&3
    & 0.956 & 4.84 & 3.88 & 19.37 & 18.91 & 3.84 & 0.567 & 3.79 \\
    \bottomrule
    \end{tabular}}
  \label{tab:ablation}%
  \vspace{-5mm}
\end{table*}%

\begin{table}[t]
  \centering
  \caption{Quantitative comparison on TGVE/TGVE+. The best results are highlighted in \textbf{bold}. $^*$ indicates retraining on our dataset.}
  
  \vspace{-3mm}
    \scalebox{.69}{
    \begin{tabular}{lcccc}
    \toprule
    \textbf{Methods}
    & \textbf{Pick Score}$\uparrow$ 
    & \textbf{TC CLIP}$\uparrow$
    & $\mathbf{ViCLIP_{\mathit{dir}}}$$\uparrow$ 
    & $\mathbf{ViCLIP_{\mathit{out}}}$$\uparrow$ 
    \\
    \midrule
    TokenFlow~\cite{geyer2023tokenflow} &
    20.58/20.62 & \textbf{0.943}/0.944 & 0.117/0.085 & 0.257/0.254 
    \\
    InsV2V~\cite{cheng2023consistentinsv2v} & 
    20.76/20.37 & 0.911/0.925 & 0.208/0.174 & 0.262/0.236 
    \\
    EVE~\cite{singer2024tgveplus}   & 
    20.76/20.88 & 0.922/0.926 & 0.221/0.198 & 0.262/0.251 
    \\
    InsV2V$^*$~\cite{cheng2023consistentinsv2v} & 
    20.83/20.91 & 0.931/0.939 & 0.237/0.191 & 0.271/0.252 
    \\
    Ours & 
    \textbf{20.90}/\textbf{20.97} & 0.941/\textbf{0.945} & \textbf{0.242}/\textbf{0.201} & \textbf{0.278}/\textbf{0.270}
    \\
    \bottomrule
    \end{tabular}}
    \label{tab:benchmark_comparison_tgve}%
    \vspace{-3mm}
\end{table}%

\noindent
\textbf{Ablation on data filtering.}
In~\cref{tab:ablation}, we investigate the role of GPT-4o~\cite{gpt4o} and Optical Flow~\cite{xu2022gmflow} (OF) filters. Removing the GPT-4o filter will decrease the performance in all three aspects, especially the textual alignment, indicating a significant loss of coherence with textual descriptions. 
Similarly, excluding the OF filter decreases all metrics while producing the most significant negative impact on temporal consistency. 
Therefore, both filters play crucial roles in maintaining coherence and motion dynamics.

\begin{figure}[t]
  \centering
   \includegraphics[width=\linewidth]{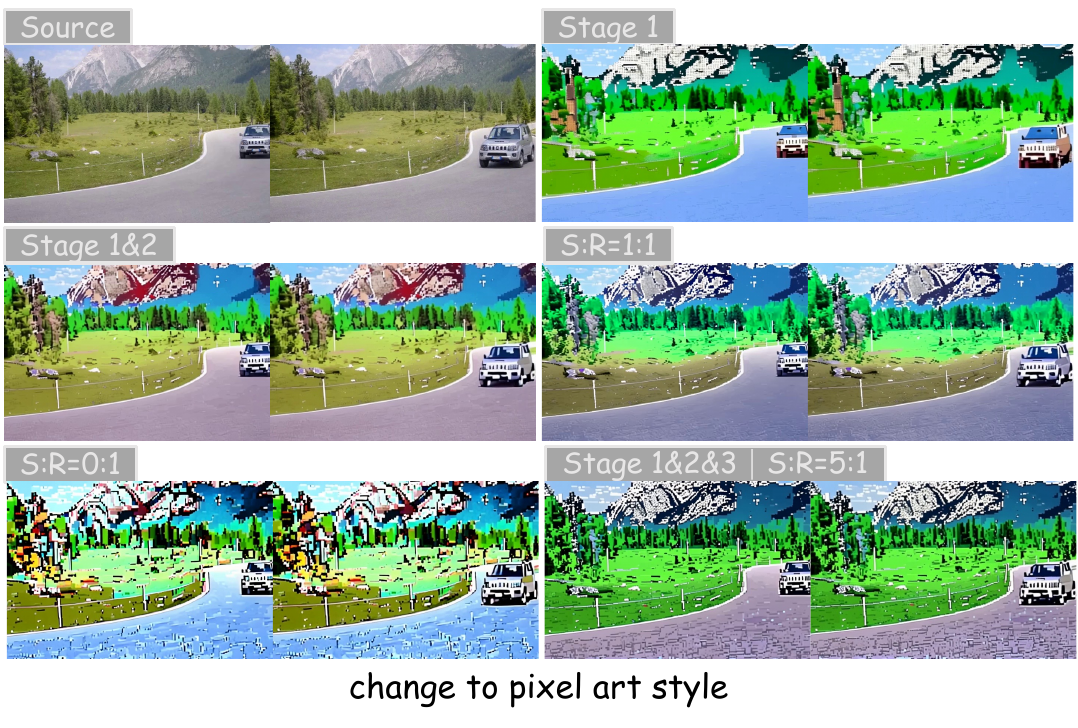}
   \setlength{\abovecaptionskip}{-0.5cm}
   \setlength{\belowcaptionskip}{-0.4cm}
   \caption{
   Visual comparisons of various ablated settings.
   }
   \label{fig:exp_abl_vis}
   \vspace{-3mm}
\end{figure}

\noindent
\textbf{Ablation on multi-stage training.}
We then evaluate the effects of multi-stage training strategy in~\cref{tab:ablation} and~\cref{fig:exp_abl_vis}. 
In ``Stage 1'', using Set-S1 yields competitive performance, while the following training stages provide further improvements. Specifically, ``Stage 1\&2'' brings considerable gain of OF EPE and textual alignment. 
In ~\cref{fig:exp_abl_vis}, ``Stage 1\&2'' is better aligned with the source video with more stylistic elements.
``Stage 1\&2\&3'' brings notable improvements in textual alignment and video quality. Compared with ``Stage 1\&2'', the DOVER and Pick scores increase favorably by 0.052 and 0.26, illustrating the benefits of LPIPS loss and high-quality static videos. The example videos in~\cref{fig:exp_abl_vis}  support the quantitative comparison.
In short, the results underscore the necessity of each training stage.

\noindent
\textbf{Ablation on static-real ratio.}
In~\cref{fig:exp_abl_vis}, ``S:R'' denotes the ratio between static and real videos. ``S:R=5:1'' provides the best alignment with the source video and instruction, offering the most satisfactory visual quality.
This ratio effectively incorporates the advantages of static and real videos, leading to the best editing ability. Thus, the 5:1 ratio was determined to be our optimal choice.

\noindent
\textbf{Ablation on the pre-trained baseline model.}
In~\cref{tab:benchmark_comparison_tgve}, we train InsV2V on our dataset (InsV2V$^*$). The improvements demonstrate the contribution of our dataset. More ablation studies can be found in the \textbf{supplementary file}.

\vspace{-2mm}
\section{Conclusion}
\label{sec:conclusion}
\vspace{-1mm}

In this paper, we proposed a large-scale instruction-based video editing dataset, InsViE-1M, which includes 1 million training triplets. 
The source data were curated from high-quality real-world videos and images, and image editing pairs. 
To ensure the quality of editing triplets, we partitioned the edited video generation process into first-frame editing and video propagation, and designed a two-stage editing-filtering construction pipeline.
In the $1^{st}$ stage, we screened the best sample from multiple edited first frames using GPT-4o to alleviate the randomness of image editing model. Then we employed GPT-4o and optical flow to evaluate each edited video and identify the satisfactory triplets in the $2^{nd}$ stage.
To achieve efficient instruction-based editing, we  introduced a multi-stage training strategy to progressively train an InsViE video editing model, whose superior video editing performance was demonstrated by our extensive experiments.

\noindent \textbf{Limitations}. Our InsViE-1M dataset and trained InsViE model have some limitations. First, our elaborately designed filtering process, though effective, is constrained by the visual understanding capability of GPT-4o. Second, the video clips in our dataset, while longer than existing video editing datasets, are still not long enough, which may limit the performance of trained models in processing longer videos. 
Third, our model performs slightly worse when adding useless content or descriptions to the input instructions because our training set includes mainly clean data.
Lastly, our InsViE model is fine-tuned on open-source video generation models, and hence may inherit their limitations in synthesizing realistic videos. 

{
    \small
    \bibliographystyle{ieeenat_fullname}
    \bibliography{main}
}

\clearpage
\setcounter{page}{1}
\maketitlesupplementary

\begin{table*}[b]
    \centering
    \vspace{-2mm}
    \caption{Examples of refined captions and instructions.}
    \begin{tabular}{@{}p{10cm}p{5.5cm}@{}}  
        \toprule
        \hspace{0.5cm} \textbf{Refined Caption} & \textbf{Instruction} \\ \midrule
        \hspace{0.5cm} The man in the blue shirt is eating a pizza on the boat. & Change the pizza to a sandwich. \\ 
        \hspace{0.5cm} The woman in the pink shirt is holding a green apple and smiling. & Replace the apple with an orange. \\ 
        \hspace{0.5cm} The red car is driving on a street with a yellow and green flag. & Make the flag blue and white. \\ 
        \hspace{0.5cm} A person holds a helmet, bright lighting highlighting its design. & Change to nighttime. \\ 
        \hspace{0.5cm} The man in the blue shirt and glasses is sitting in a room. & Add snow effect to the room. \\ 
        \hspace{0.5cm} The man in the gray jacket is driving a car. & Convert to watercolor portrait. \\ 
        \bottomrule
    \end{tabular}
    \vspace{-2mm}
    \label{tab:supp_example_modifications}
\end{table*}

In this supplementary file, we provide additional details of the construction pipeline of our InsViE-1M dataset in~\cref{sec:supp_insvie1m}, additional settings of model training and testing in~\cref{sec:supp_trianing_testing}, more visual comparisons in~\cref{sec:supp_visual_results}, and more ablation studies in~\cref{sec:supp_ablation_study}.
In addition, we provide a demo video that includes more visual comparisons. Please view the video using software that can open MOV files. 

\section{Details of InsViE-1M Dataset}
\label{sec:supp_insvie1m}
In this section, we first show the specific prompts used for generating instruction and filtering in ~\cref{subsec:supp_prompt_recap_instructuion} and ~\cref{subsec:supp_prompt_screen_filter}, respectively. Then we illustrate the case study on CFG in~\cref{subsec:supp_CFGs}. Finally, we present examples of the data construction process in~\cref{subsec:supp_eg_triplets}.

\subsection{Prompts for Recaptioning and Instruction Generation}
\label{subsec:supp_prompt_recap_instructuion}
 
\begin{figure}[t]
  \centering
   \includegraphics[width=\linewidth]{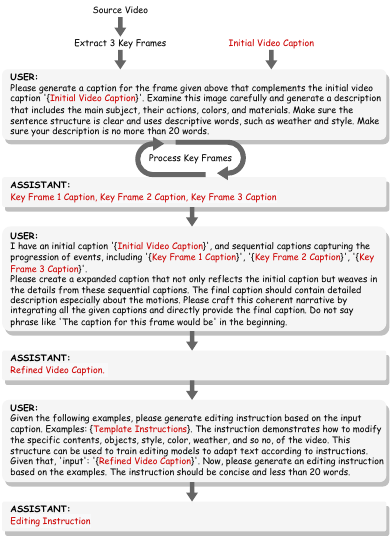}
   \setlength{\abovecaptionskip}{-0.1cm}
   \setlength{\belowcaptionskip}{-0.3cm}
   \vspace{-3mm}
   \caption{
    Pipeline and prompts of recaptioning and instruction generation.
   }
   \label{fig:supp_prompt_recap_instruction}
   \vspace{-4mm}
\end{figure}

The original video dataset provides initial video captions that outline the overall content of the videos. However, these captions are often either too long, containing excessive details, or too brief, consisting of only a few words. As a result, they are not suitable for generating effective instructions.
Therefore, as shown in~\cref{fig:supp_prompt_recap_instruction}, we propose a systematic approach to generate video captions and the corresponding editing instructions by a large vision-language model~\cite{hong2024cogagent}.
The process begins with extracting three key frames from the source video to capture important moments. 
Based on the initial captions, the system generates supplementary descriptions for each key frame, capturing the actions and nuances within the frames. This ensures a coherent narrative that aligns with the initial caption while highlighting the key elements of the scene.
Then we use the initial caption and the generated key frame captions to produce the final video caption.
Finally, based on user-provided examples like~\cite{brooks2023instructpix2pix}, the system generates concise editing instructions from the final caption. These instructions guide the editing of video content, including specific objects, styles, colors, and weather conditions, enabling adaptive adjustments to the captions. Through this process, we effectively produce high-quality video captions and flexible editing instructions.

In~\cref{tab:supp_example_modifications}, we list the refined caption samples and the corresponding instructions with different editing types.

\subsection{Prompts for Screening and Filtering}
\label{subsec:supp_prompt_screen_filter}
We illustrate the screening and filtering process mentioned in Sec. 3 of the main paper, and present the simplified prompts in Fig. 2 of the main paper. Below, we provide the complete prompts along with the input format for the GPT-4o API~\cite{gpt4o}. The prompts for screening are first presented, followed by the prompts employed for filtering.

\begin{mdframed}[linecolor=black, linewidth=1pt]
\textbf{Prompts of Screening:}

\noindent
\textbf{System:}

\noindent
You are an advanced AI model specifically trained to assess the naturalness of edited images. Your task is to evaluate a set of edited images based on their adherence to the original image and the provided editing instructions. Here’s how to perform the evaluation:

\noindent
- Strict Adherence: Assess whether each edited image strictly follows the provided instructions. The modifications should directly reflect the requested changes without any deviations.

\noindent
- Integration of Edits: Assess whether each edited image is seamlessly blended with the original image. The modifications should maintain a visual balance and consistency in color and tone.

\noindent
- Absence of Artifacts: Evaluate whether the edits appear natural and free from any noticeable artifacts that would detract from the image.

\noindent
- Subject Matter Consistency: Check for any distortions or elements that could have been introduced during the editing process. The edited images should be consistent in terms of lighting and shadows.

\noindent
- Identify the Best Edit: Determine which edited image best reflects the requested changes and appears the most natural compared to the original.

\noindent
\textbf{User}:

\noindent
Please evaluate the following images based on their quality and natural appearance:
The first image is the original image, and the next five images are the edited images. Editing Instructions: \{instruction\}. 
Based on your evaluation, identify which edited image best adheres to the original and editing instructions. Specify which image it is (0 through 5).
Return the result as a Python dictionary string with the key `best\_image' indicating the number of the best image.
DO NOT PROVIDE ANY OTHER OUTPUT TEXT OR EXPLANATION. Only provide the Python dictionary string.
For example, your response should look like this: \{`best\_image': 3\}.

\noindent
This is the first image: \{`source\_url'\}.

\noindent
Edited images are as follows: \{`edited\_url\_0'\}, \{`edited\_url\_1'\}, \{`edited\_url\_2'\}, \{`edited\_url\_3'\}, \{`edited\_url\_4'\}, \{`edited\_url\_5'\}.
\end{mdframed}

\begin{mdframed}[linecolor=black, linewidth=1pt]
\textbf{Prompts of Filtering:}

\noindent
\textbf{System:}

\noindent
You are an advanced AI tasked with evaluating the quality of video edits based on the adherence to specific editing instructions and the consistency of the edited frames. Your evaluation should focus on the following criteria:

\noindent
- Strict Adherence: Assess whether each edited image strictly follows the provided instructions. The modifications should directly reflect the requested changes without any deviations.

\noindent
- Integration of Edits: Assess whether each edited image is seamlessly blended with the original image. The modifications should maintain a visual balance and consistency in color and tone.

\noindent
- Absence of Artifacts: Evaluate whether the edits appear natural and free from any noticeable artifacts that would detract from the image.

\noindent
- Subject Matter Consistency: Check for any distortions or elements that could have been introduced during the editing process. The edited images should be consistent in terms of lighting and shadows.

\noindent
- Composition Coherence: Examine the overall composition after the edits. The layout should maintain the visual balance across the frames.

\noindent
- Content Consistency: Compare the edited frames with the original frames, ensuring that the contents are consistent across the frames.

\noindent
Please conduct this evaluation by meticulously applying these criteria to determine the quality of the edits.

\noindent
\textbf{User}:

\noindent
Please evaluate the following video edit based on the provided instructions:
The first three frames are from the original video, and the last three frames are from the edited video. Editing Instructions: \{instruction\}
Based on your evaluation, answer the following questions:
(1) Provide your evaluation solely as a quality score where the quality score is an integer value between 1 and 5, with 5 indicating the highest level of adherence to the instructions and overall quality.
(2) Describe the aspects of the edit that were not executed well, including any artifacts or inconsistencies detected.
Please generate the response in the form of a Python dictionary string with key `score'. `score' should be an integer indicating the quality score.
DO NOT PROVIDE ANY OTHER OUTPUT TEXT OR EXPLANATION.
For example, your response should look like this: \{`score': 3\}.

\noindent
Images from source video: \{`source\_url\_0'\}, \{`source\_url\_1'\}, \{`source\_url\_2'\}.

\noindent
Images from edited video: \{`edited\_url\_0'\}, \{`edited\_url\_1'\}, \{`edited\_url\_2'\}.

\end{mdframed}

\begin{figure*}[ht]
 \centering
  \includegraphics[width=\linewidth]{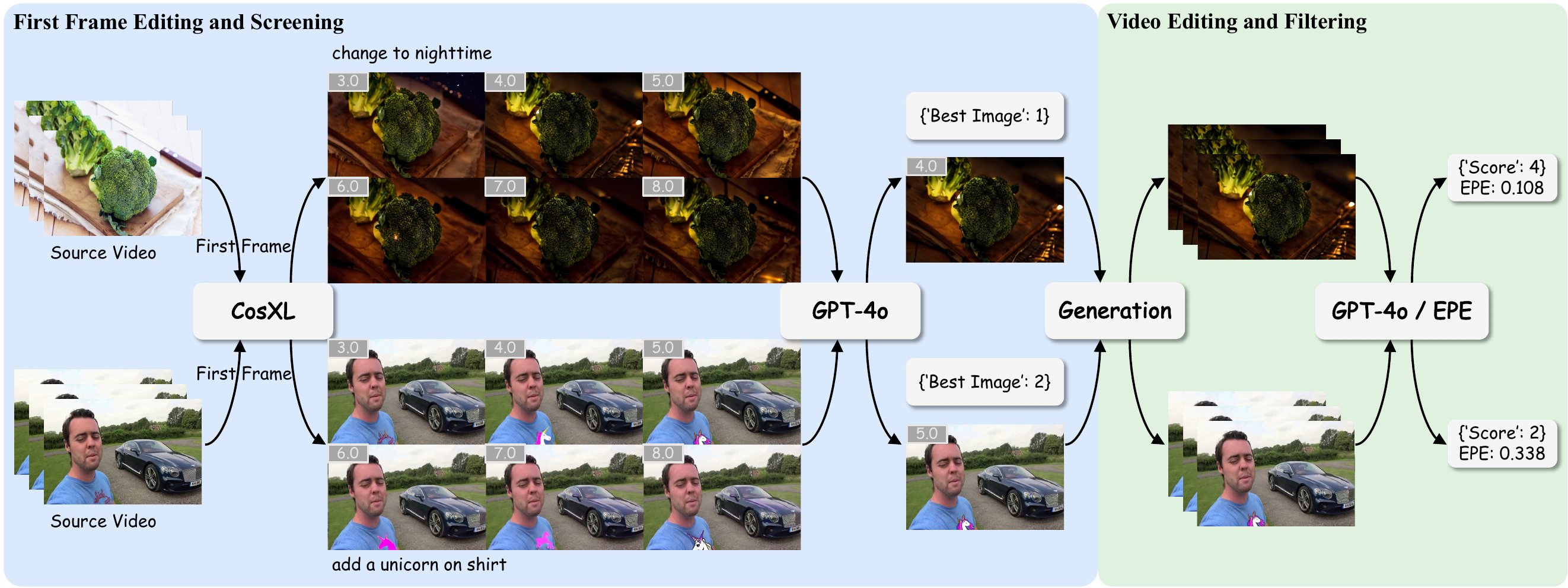}
   \caption{
   Examples of generating triplets from real-world videos.}
   \label{fig:supp_triplet_0}
\end{figure*}
\begin{figure*}[ht]
 \centering
  \includegraphics[width=\linewidth]{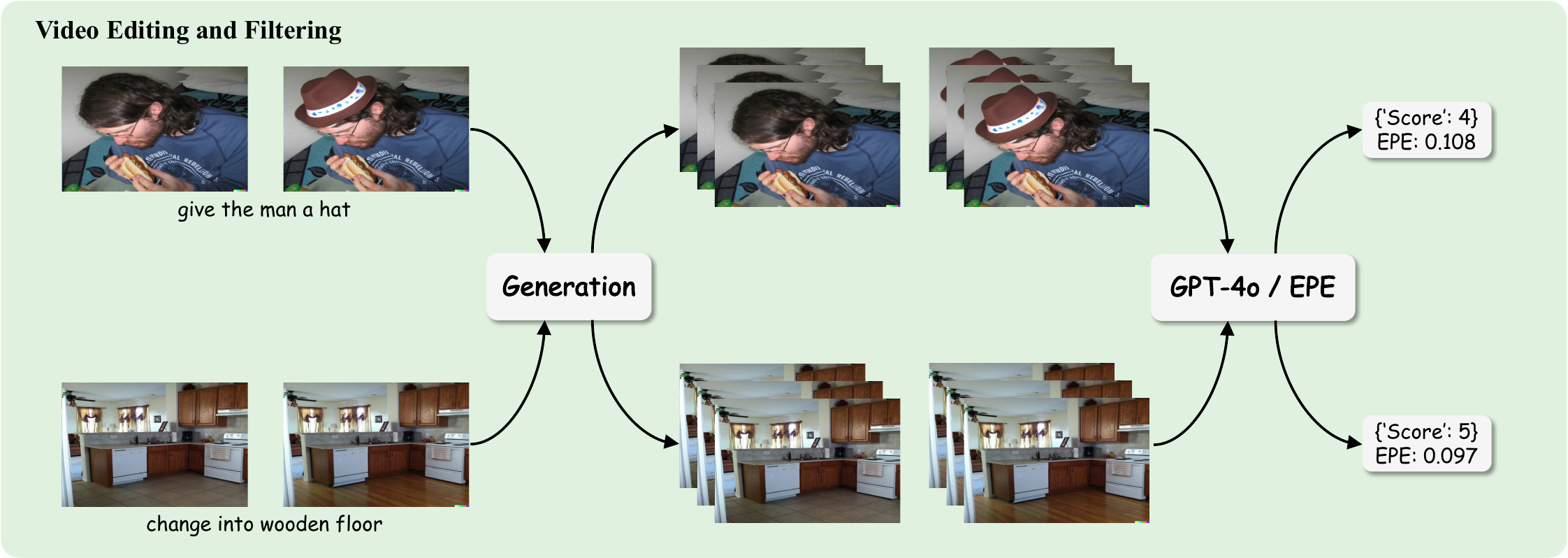}
   \caption{
   Examples of generating triplets from image editing pairs.}
   \label{fig:supp_triplet_1}
\end{figure*}

\begin{figure*}[ht]
 \centering
  \includegraphics[width=\linewidth]{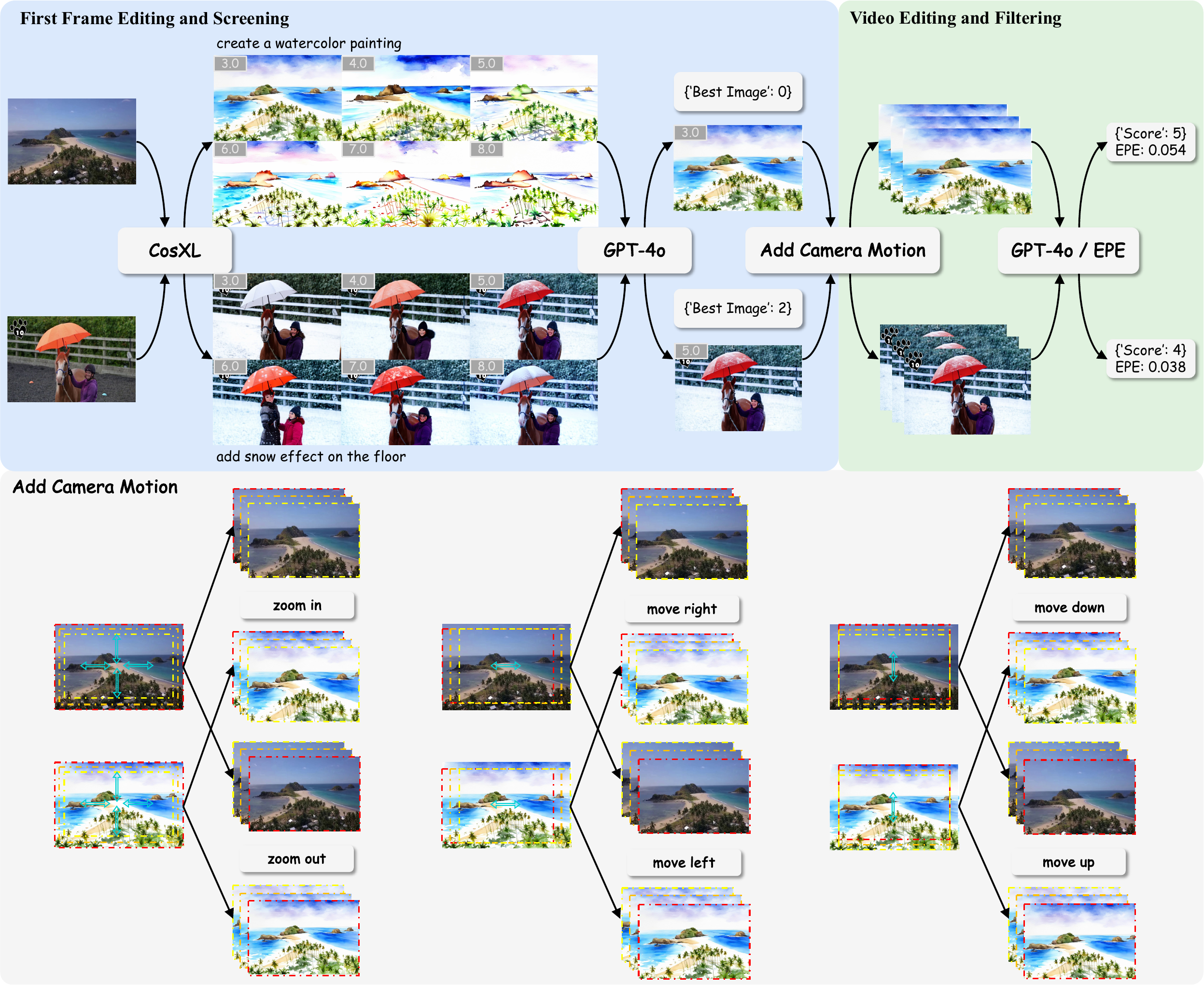}
   \caption{
   Examples of generating triplets from real-world images.}
   \label{fig:supp_triplet_2}
\end{figure*}


\subsection{Examples of Triplets Construction Process}
\label{subsec:supp_eg_triplets}
In this section, we provide examples of the construction process of the training triplets. 

\noindent
\textbf{Triplet generation from real-world videos.}
In~\cref{fig:supp_triplet_0}, we show two examples of the triplet generation from real-world videos by simplifying the intermediate process.

\noindent
\textbf{Triplet generation from image editing pairs.}
In~\cref{fig:supp_triplet_1}, we show two examples of the triplet generation from image editing pairs by simplifying the intermediate process.

\noindent
\textbf{Generate static video triplets from real-world images.}
In~\cref{fig:supp_triplet_2}, we show two examples of the triplet generation from real-world images by simplifying the intermediate process. Most of the construction pipeline is the same with triplet generation from real-world videos, while the generation step in ``Video Editing and Filtering'' is replaced by the addition of the camera motion.
Specifically, we illustrate the detailed process of adding camera motion in the bottom example of~\cref{fig:supp_triplet_2}, which is mentioned in Sec. 3.3 of the main paper.
For ``zoom in'' and ``zoom out'', we set the minimum cropping size to 90\% of the original image size and produce image sequences by gradually decreasing or increasing the cropping size.
For ``move right'', ``move left'', ``move down'' and ``move up'', we set the cropping size to 90\% of the original image size and produce image sequences by gradually adjusting the cropping location.

\subsection{The Selection of CFGs}
\label{subsec:supp_CFGs}
\begin{figure}[ht]
 \centering
  \includegraphics[width=\linewidth]{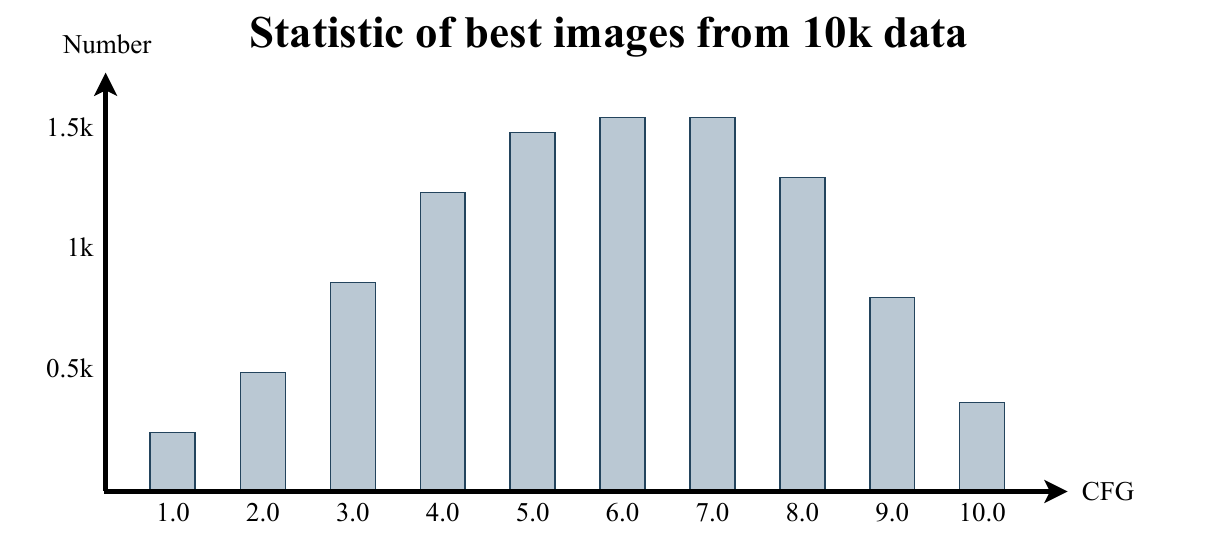}
   \setlength{\abovecaptionskip}{-0.3cm}
   \setlength{\belowcaptionskip}{-0.5cm}
   \caption{
   Statistic of the best edited images with different CFGs on 10K images.   }
   \label{fig:supp_CFGs}
\end{figure}

In Sec. 3.1 of the main paper, we choose a range CFGs to edit the video first frames. We randomly select 10K first frames and images from our initial dataset and utilize CosXL~\cite{cosxl} to produce the edited outputs using various CFGs (from 1.0 to 10.0) for each image.
Then we use GPT-4o~\cite{gpt4o} to screen the edited images and count the numbers of best edited images produced by each CFG. 
As shown in~\cref{fig:supp_CFGs}, most of the best samples can be generated with CFGs from 3.0 to 8.0. 
Therefore, we set CFG within [3, 8] to generate 6 edited samples, which is also acceptable in terms of resource consumption.

\section{Training and Testing}
\label{sec:supp_trianing_testing}

\textbf{Implementation details.} 
We train the InsViE model using similar settings to the default settings of CogVideoX~\cite{yang2024cogvideox}. The training is conducted on 8 nodes, each equipped with 8 Nvidia A100 GPUs, utilizing a batch size of 128 for a total of 40k steps. The Adam optimizer is employed with exponential moving average (EMA), setting the learning rate to 1e-3, betas to 0.9 and 0.95, weight decay to 1e-5, and EMA decay to 0.9999. The training data comprises a diverse set of video samples, with 720$\times$480 pixels and 25 frames per video, ensuring consistency across inputs.
At the last stage, both the weight of LPIPS loss and $L_2$ loss are set as 0.5. 

\noindent \textbf{Prompt for GPT score of testing.}
In terms of using GPT-4o to evaluate the edited videos, the selection of frames and prompts differs from the screening and filtering process outlined in data construction pipeline. 
Firstly, instead of sampling three key frames from video pairs as described in Sec. 3.1 of the main paper, we input all the frames to GPT-4o in testing stage, since the resource consumption associated with the scale of the test set is acceptable. 
Secondly, according to the ``Evaluation Metrics'' in Sec. 5.1 of the main paper, we provide the scores of GPT-4o across three aspects as a new metric. 
The original prompts used in the screening and filtering process are slightly adjusted.
To be specific, the prompts for evaluating temporal consistency and textual alignment are modified to concentrate on each respective aspect, while the prompts for evaluating the video quality remains the same as the prompts of filtering. The revised prompts are shown below.

\begin{mdframed}[linecolor=black, linewidth=1pt, nobreak]
\textbf{Temporal Consistency:}

\noindent
\textbf{System:}

\noindent
You are an advanced AI tasked with evaluating the quality of video edits based on the adherence to specific editing instructions and the consistency of the edited frames. Your evaluation should focus on the following criteria:

\noindent
- Composition Coherence: Examine the overall composition after the edits. The layout should maintain the visual balance across the frames.

\noindent
- Content Consistency: Compare the edited frames with the original frames, ensuring that the contents are consistent across the frames.

\noindent
Please conduct this evaluation by meticulously applying these criteria to determine the quality of the edits.

\noindent
\textbf{User}:

\noindent
Please evaluate the following video edit based on the provided instructions:
The first half of the frames are from the original video, and the second half of the frames are from the edited video. Editing Instructions: \{instruction\}
Based on your evaluation, answer the following question:
Provide your evaluation solely as a score that is an integer value between 1 and 5, with 5 indicating the highest level of temporal consistency between videos and across the frames.
Please generate the response in the form of a Python dictionary string with key `score'. `score' should be an integer indicating the temporal consistency score.
DO NOT PROVIDE ANY OTHER OUTPUT TEXT OR EXPLANATION.
For example, your response should look like this: \{`score': 3\}.

\noindent
Images from source video: \{`source\_url\_0'\}, ..., \{`source\_url\_n'\}.

\noindent
Images from edited video: \{`edited\_url\_0'\}, ..., \{`edited\_url\_n'\}.

\end{mdframed}

\begin{mdframed}[linecolor=black, linewidth=1pt, nobreak]
\textbf{Textual Alignment:}

\noindent
\textbf{System:}

\noindent
You are an advanced AI tasked with evaluating the quality of video edits based on the adherence to specific editing instructions and the consistency of the edited frames. Your evaluation should focus on the following criteria:

\noindent
- Strict Adherence: Assess whether each edited image strictly follows the provided instructions. The modifications should directly reflect the requested changes without any deviations.


\noindent
- Integration of Edits: Assess whether each edited image is seamlessly blended with the original image. The modifications should maintain a visual balance and consistency in color and tone.

\noindent
\textbf{User}:

\noindent
Please evaluate the following video edit based on the provided instructions:
The first half of the frames are from the original video, and the second half of the frames are from the edited video. Editing Instructions: \{instruction\}
Based on your evaluation, answer the following question:
Provide your evaluation solely as a score that is an integer value between 1 and 5, with 5 indicating the highest level of textual alignment of the edited video frames.
Please generate the response in the form of a Python dictionary string with key `score'. `score' should be an integer indicating the textual alignment score.
DO NOT PROVIDE ANY OTHER OUTPUT TEXT OR EXPLANATION.
For example, your response should look like this: \{`score': 3\}.

\noindent
Images from source video: \{`source\_url\_0'\}, ..., \{`source\_url\_n'\}.

\noindent
Images from edited video: \{`edited\_url\_0'\}, ..., \{`edited\_url\_n'\}.

\end{mdframed}

\section{More Visual Results}
\label{sec:supp_visual_results}
\begin{figure*}[ht]
 \centering
  \includegraphics[width=\linewidth]{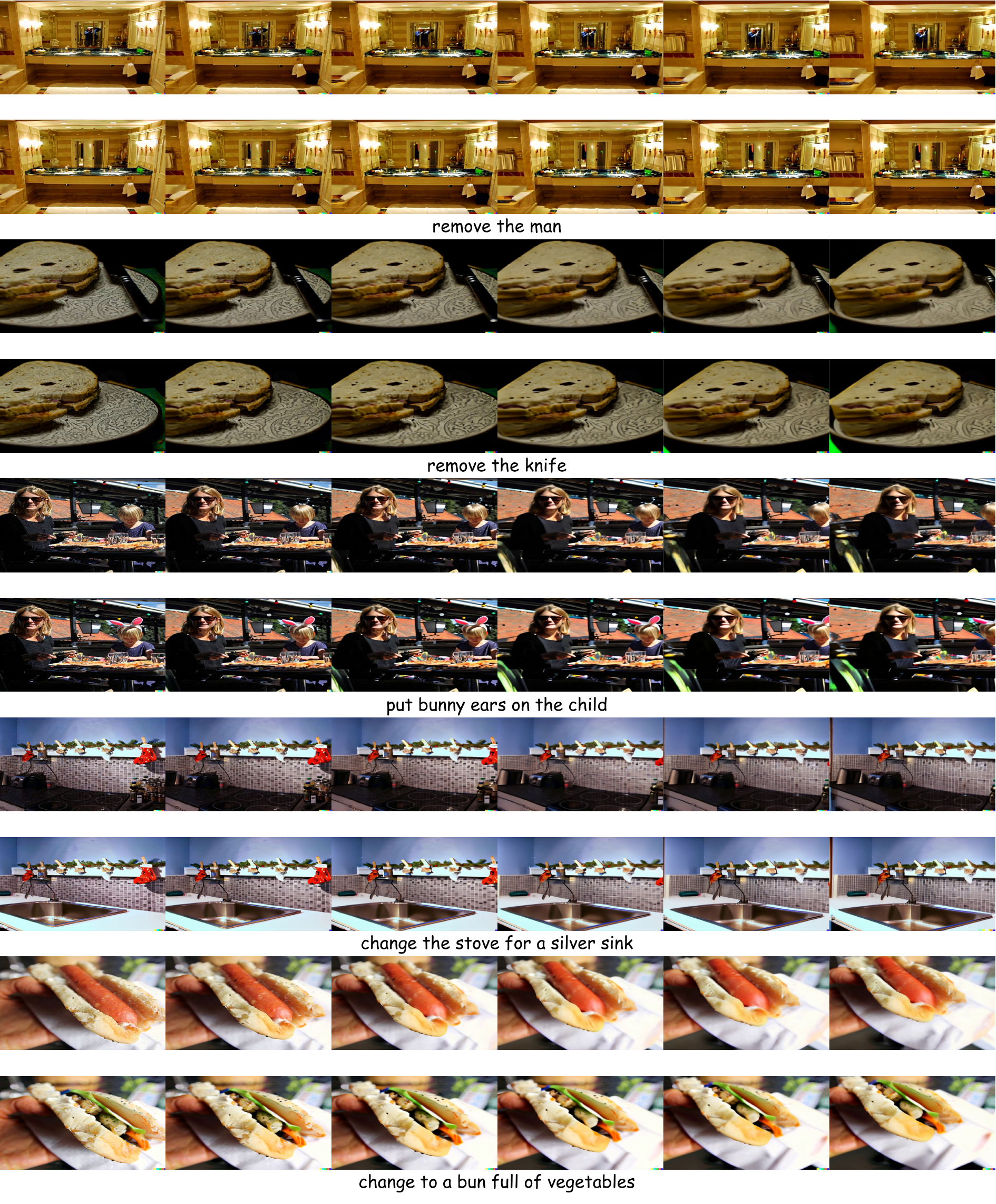}
   \caption{
   Sample triplets of our InsViE-1M dataset. For each sample, from top to bottom: original video, edited video, instruction.}
   \label{fig:supp_dataset_0}
\end{figure*}

\begin{figure*}[ht]
 \centering
  \includegraphics[width=\linewidth]{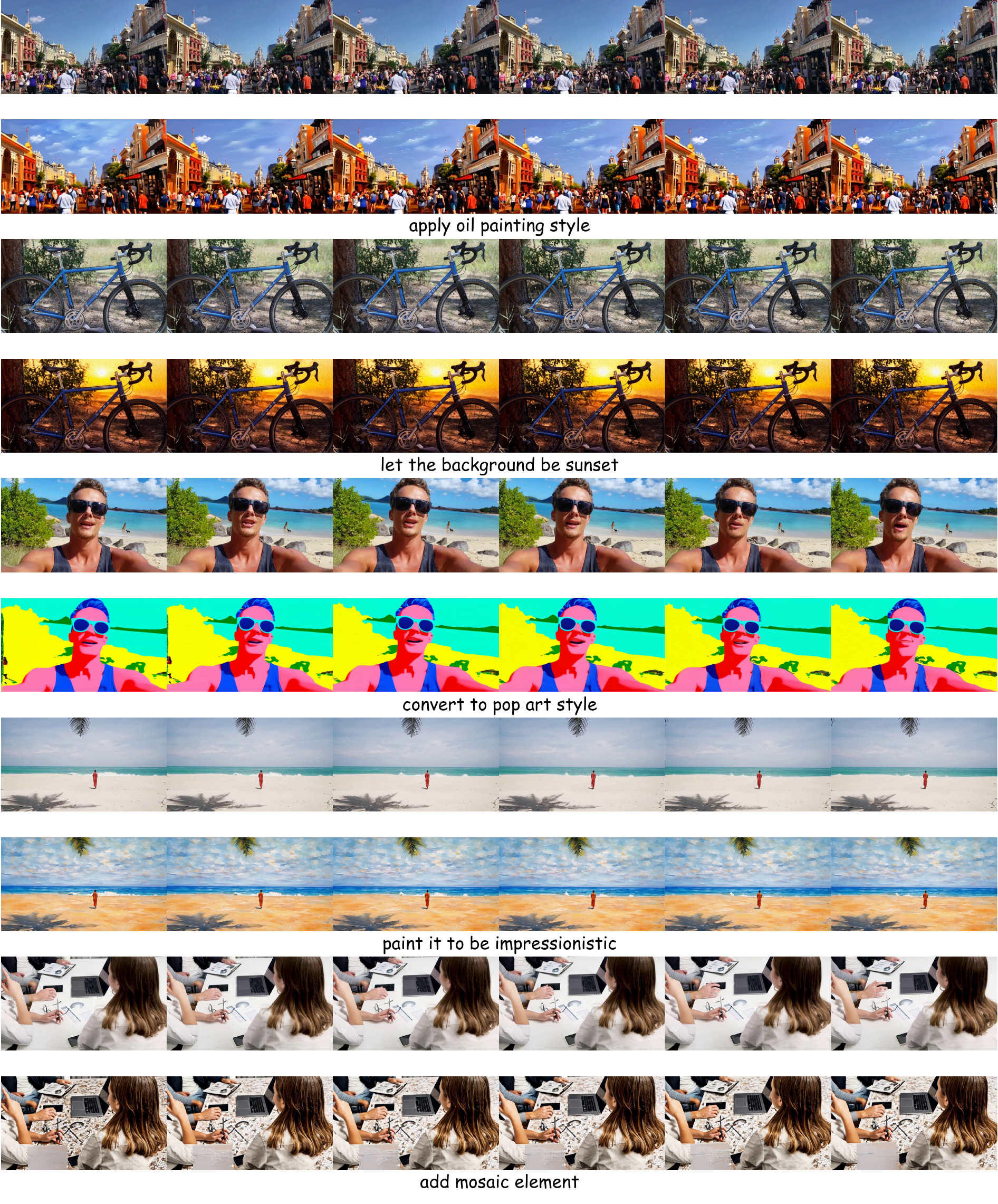}
   \caption{
   Sample triplets of our InsViE-1M dataset. For each sample, from top to bottom: original video, edited video, instruction.}
   \label{fig:supp_dataset_3}
\end{figure*}

\begin{figure*}[ht]
 \centering
  \includegraphics[width=\linewidth]{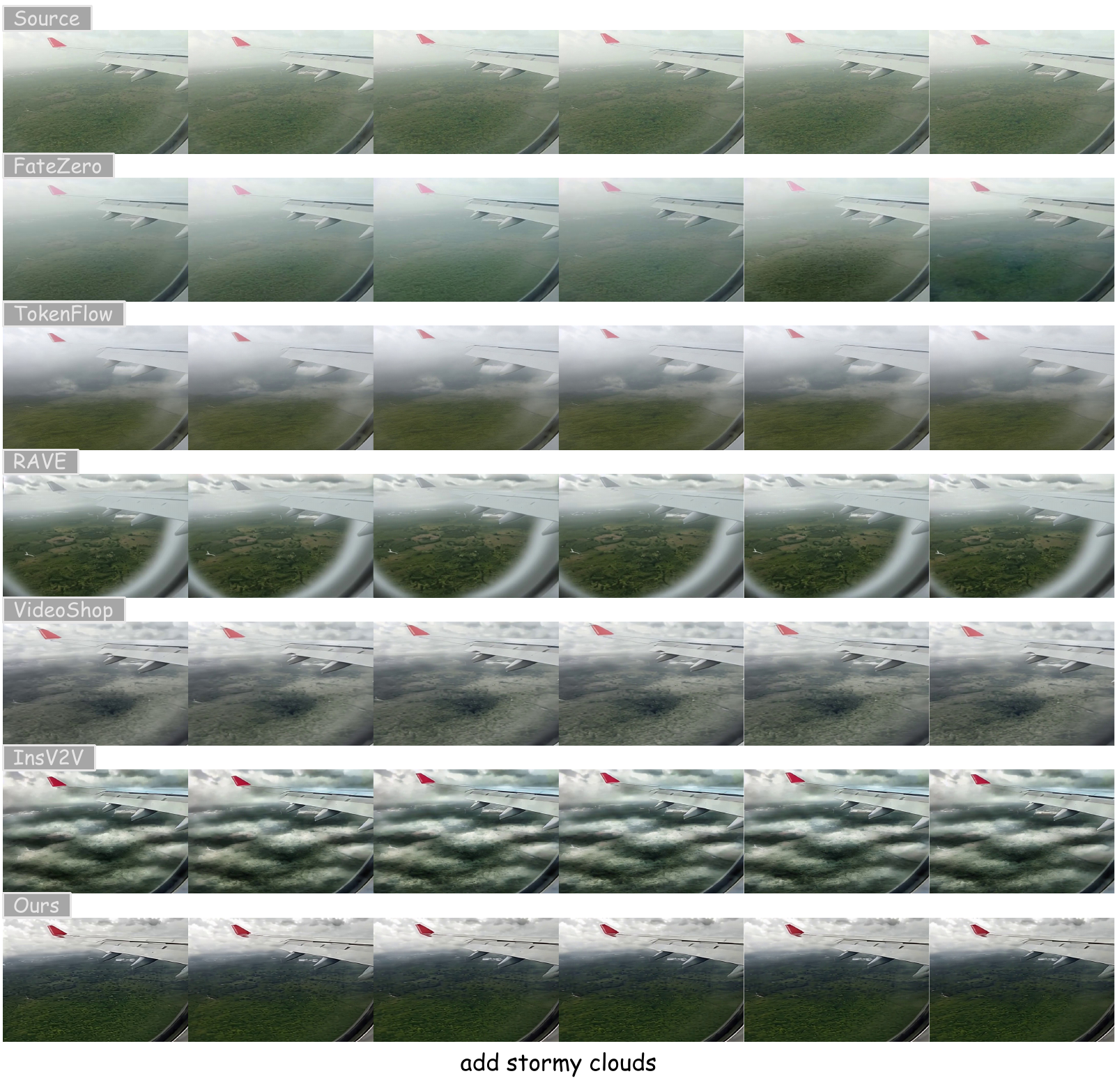}
   \caption{
   Visual comparison between our InsViE model and state-of-the-art methods.}
   \label{fig:supp_results_4}
\end{figure*}

\begin{figure*}[ht]
 \centering
  \includegraphics[width=\linewidth]{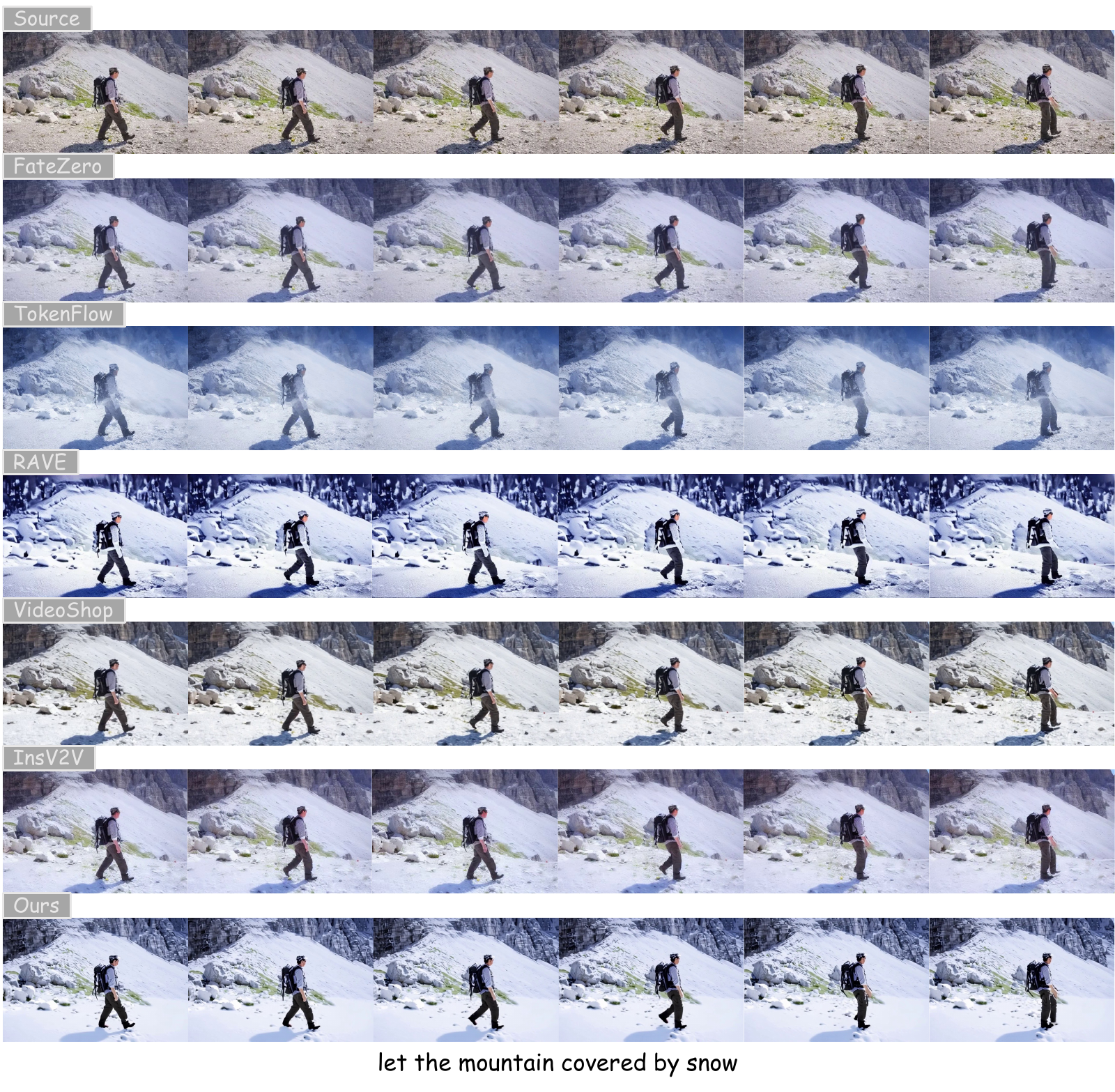}
   \caption{
   Visual comparison between our InsViE model and state-of-the-art methods.}
   \label{fig:supp_results_6}
\end{figure*}

\begin{figure*}[ht]
 \centering
  \includegraphics[width=\linewidth]{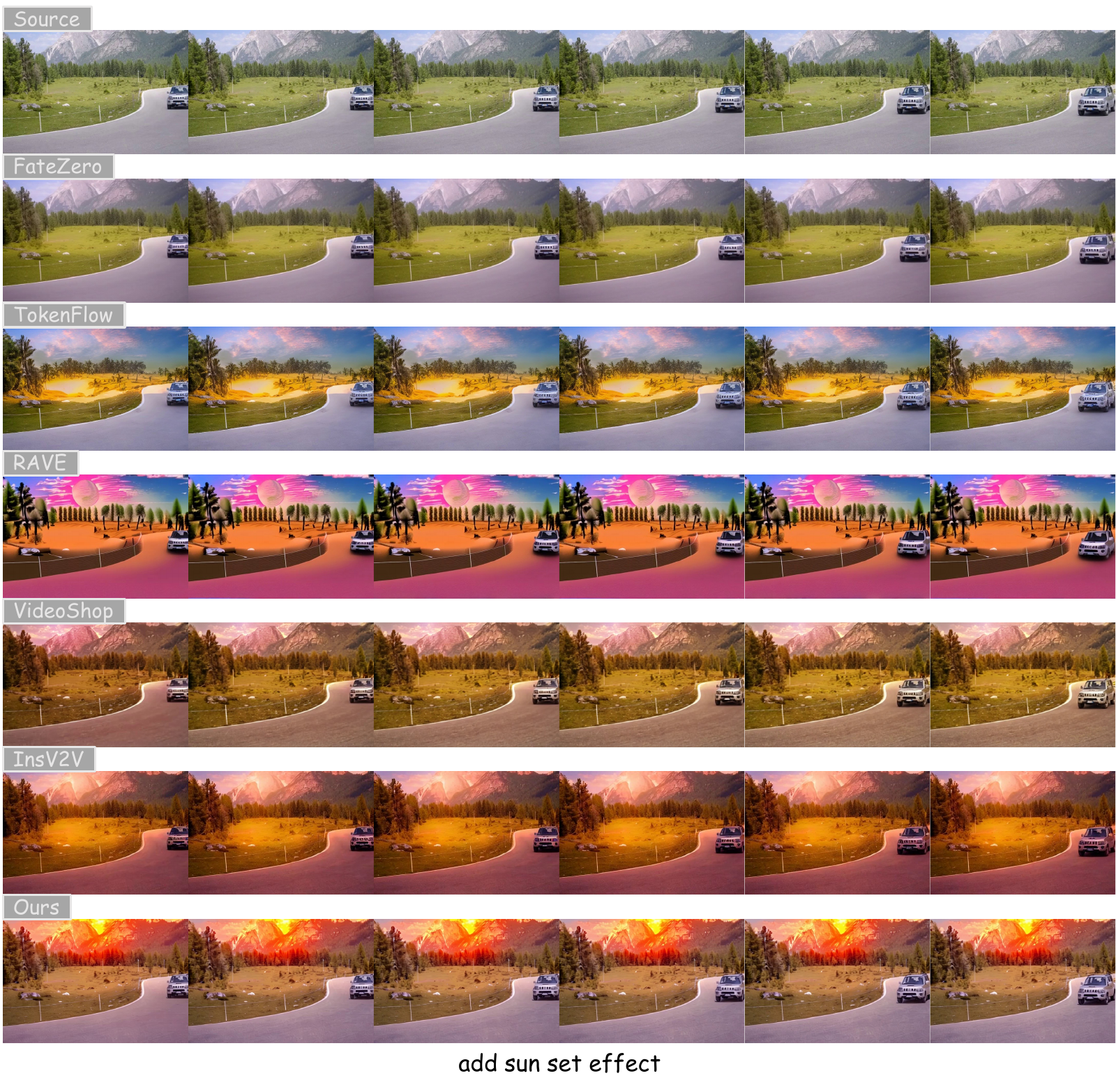}
   \caption{
   Visual comparison between our InsViE model and state-of-the-art methods.}
   \label{fig:supp_results_7}
\end{figure*}

\begin{figure*}[ht]
 \centering
  \includegraphics[width=\linewidth]{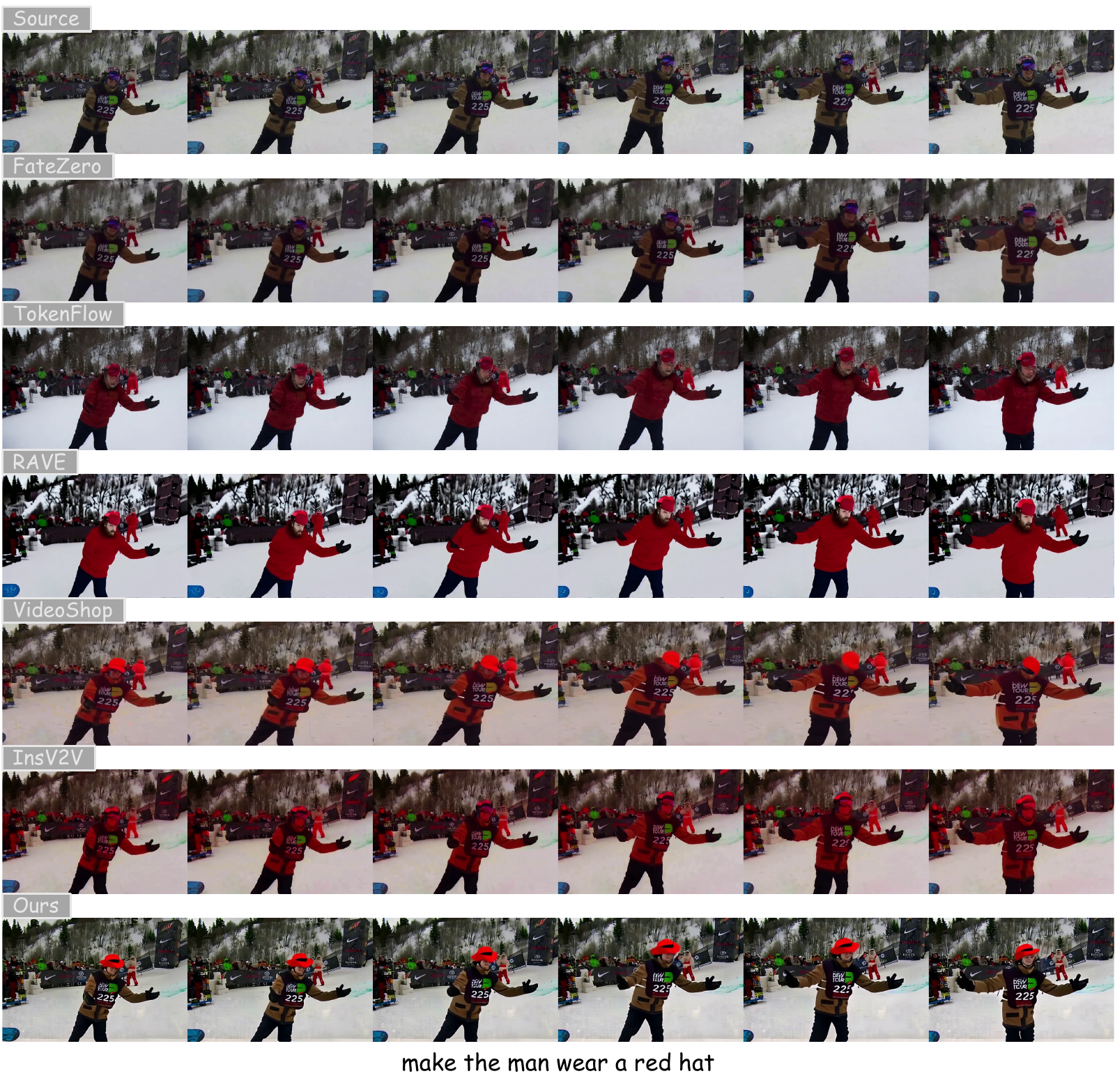}
   \caption{
   Visual comparison between our InsViE model and state-of-the-art methods.}
   \label{fig:supp_results_8}
\end{figure*}

\begin{figure*}[ht]
 \centering
  \includegraphics[width=\linewidth]{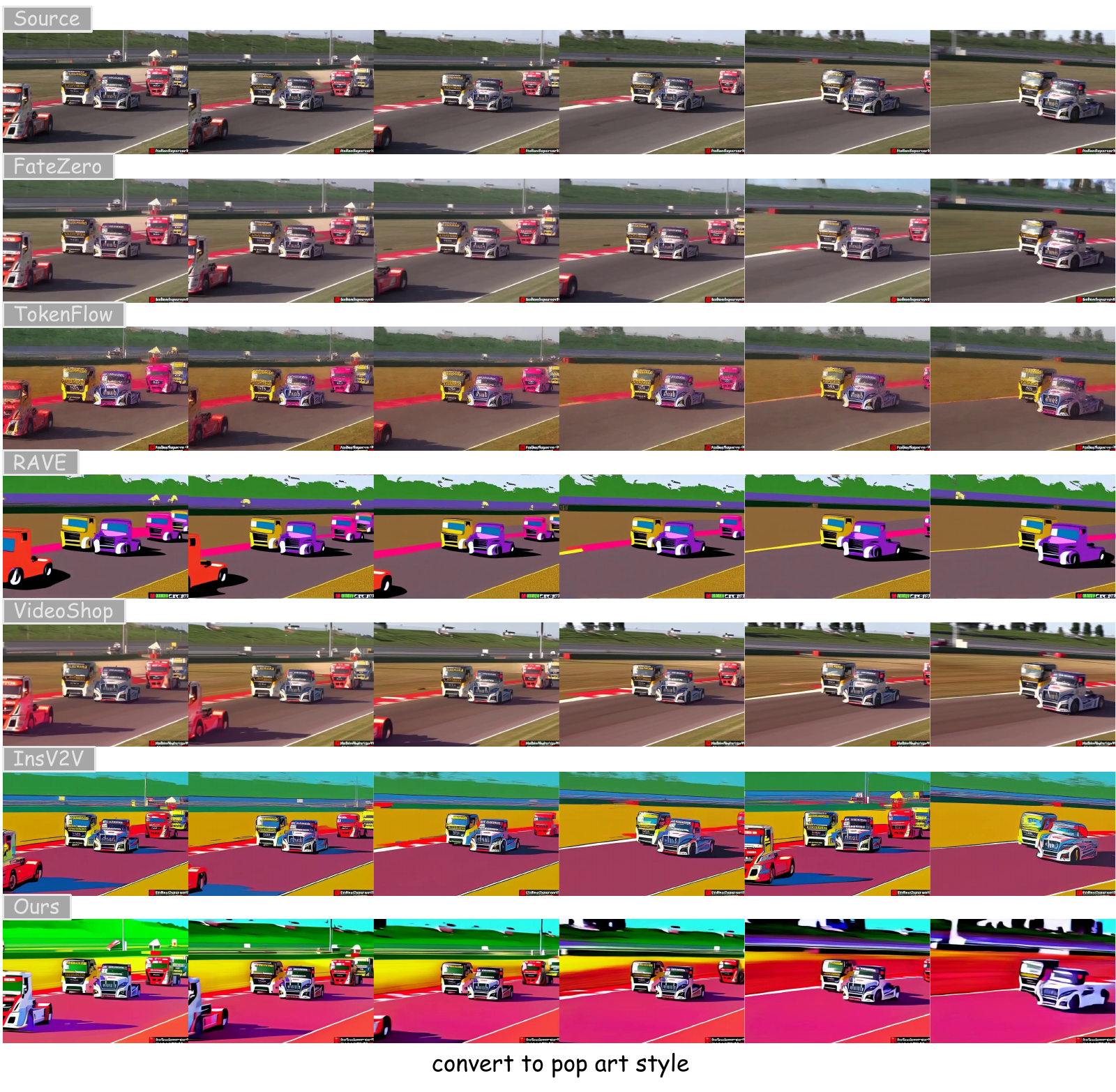}
   \caption{
   Visual comparison between our InsViE model and state-of-the-art methods.}
   \label{fig:supp_results_9}
\end{figure*}

In this section, we provide more samples of InsViE-1M dataset and more qualitative comparisons between InsViE and previous methods.
As shown in~\cref{fig:supp_dataset_0,fig:supp_dataset_3}, we present more triplet samples of our InsViE-1M dataset, including removal, substitution, addition, stylization, \etal.
Additional comparisons with previous methods are shown in~\cref{fig:supp_results_4,fig:supp_results_6,fig:supp_results_7,fig:supp_results_8,fig:supp_results_9}.
From the visual comparisons, one can see that our InsViE model achieves better editing performance among various editing instructions, producing more visually more pleasing videos.

\begin{table*}[t]
  \centering
  \caption{Ablation study on static-real ratio in the final training stage.
  }
    \scalebox{.89}{
    \begin{tabular}{c||c|c|c||c|c|c||c|c}
    \toprule
    \multirow{2}[2]{*}{\textbf{Training Settings}} 
    & \multicolumn{3}{c||}{\textbf{Temporal Consistency}} 
    & \multicolumn{3}{c||}{\textbf{Textual Alignment}} 
    & \multicolumn{2}{c}{\textbf{Video Quality}}
    \\
    \cmidrule{2-9} 
    & \textbf{CLIP} $\uparrow$ & \textbf{OF EPE} $\downarrow$ & \textbf{GPT Score} $\uparrow$
    & \textbf{CLIP} $\uparrow$ & \textbf{Pick Score} $\uparrow$ & \textbf{GPT Score} $\uparrow$ 
    & \textbf{DOVER} $\uparrow$ & \textbf{GPT Score} $\uparrow$
    \\
    \midrule
    Static:Real=0:1
    & 0.951 & 4.88 & 3.82 & 19.15 & 18.70 & 3.79 & 0.519 & 3.65 \\
    Static:Real=0.5:1
    & 0.954 & 4.89 & 3.86 & 19.21 & 18.73 & 3.80 & 0.540 & 3.71 \\
    Static:Real=1:1
    & 0.956 & 4.85 & 3.87 & 19.18 & 18.69 & 3.79 & 0.547 & 3.72 \\
    Static:Real=5:1
    & 0.956 & 4.84 & 3.87 & 19.37 & 18.91 & 3.84 & 0.567 & 3.79 \\
    \bottomrule
    \end{tabular}}
  \label{tab:ablation}%
\end{table*}%

\section{More Ablation Studies}
\label{sec:supp_ablation_study}

\begin{table}[t]
  \centering
    \caption{Ablation study on the LPIPS loss in Stage 3.}
    \scalebox{.8}{
    \begin{tabular}{lcccc}
    \toprule
    \textbf{Stage 1\&2\&3} 
    & \textbf{TC GPT}$\uparrow$ 
    & \textbf{TA GPT}$\uparrow$ 
    & \textbf{DOVER}$\uparrow$ 
    & \textbf{VQ GPT}$\uparrow$ \\
    \midrule
    w/ LPIPS & \textbf{3.88} & \textbf{3.84} & \textbf{0.567} & \textbf{3.79} \\
    w/o LPIPS & 3.86 & 3.83 & 0.543 & 3.73 \\
    \bottomrule
    \end{tabular}}
    \label{tab:ablation_lpips}
\end{table}

\noindent
\textbf{Ablation on the LPIPS loss in Stage 3.}
As described in Sec. 4.2 in the main paper, we use $L_2$ loss in the first two training stages. LPIPS loss is added in the final stage to enhance detail generation. As shown in \cref{tab:ablation_lpips}, it contributes more to video quality metrics.

\noindent
\textbf{Ablation on static-real ratio.}
We further investigate the impact of different ratios of static to real videos in Set-S3. 
In~\cref{tab:ablation}, ``Static:Real=0:1'' exhibits similar results to ``Stage 1\&2'', indicating the limitation of using real videos only.
Increasing the ratio to ``0.5:1'' leads to better results than ``Stage 1\&2'' on all the metrics.
By setting ``Static:Real=1:1'', the model's performance stabilizes with better DOVER and GPT quality scores, demonstrating the benefits of static videos for visual quality. 
The most notable gain can be observed at ``Static:Real=5:1'', especially on the textual alignment and video quality. 

\end{document}